\pdfoutput=1

\documentclass[11pt]{article}

\usepackage{acl}

\usepackage{times}
\usepackage{latexsym}

\usepackage[T1]{fontenc}

\usepackage[utf8]{inputenc}

\usepackage{microtype}

\usepackage{inconsolata}
\usepackage{microtype}
\usepackage{graphicx}
\usepackage{import}
\usepackage{layout}
\usepackage{tabularx, makecell}
\usepackage{booktabs}
\usepackage{mathrsfs}
\usepackage{amssymb} 
\usepackage{url}
\usepackage{hyperref}
\usepackage{graphicx}
\usepackage{xspace,paralist}
\usepackage{times,latexsym}
\usepackage{amsmath}
\usepackage{appendix}
\usepackage{comment} 
\usepackage{enumitem}
\usepackage{makecell}
\usepackage{multirow}
\usepackage{xcolor}
\usepackage{arydshln}
\usepackage{cleveref}
\usepackage{todonotes}
\usepackage{longtable,supertabular}
\usepackage{amssymb}
\usepackage{pifont}
\newcommand{\cmark}{\ding{51}}%
\newcommand{\xmark}{\ding{55}}%

\newcommand{\tabref}[2][]{Table#1~\ref{#2}\xspace}
\newcommand{\figref}[1]{Figure~\ref{#1}\xspace}
\newcommand{\secref}[1]{Section~\ref{#1}\xspace}
\newcommand{\appref}[1]{Appendix~\ref{#1}\xspace}

\newcommand{\model}[1]{\text{#1}\xspace}
\newcommand{\factscore}{\model{FactScore}}
\newcommand{\factool}{\model{FacTool}}
\newcommand{\factor}{\model{FACTOR}}
\newcommand{\rarr}{\model{RARR}}
\newcommand{\cove}{\model{CoVe}}
\newcommand{\perplexityai}{\model{Perplexity.ai}}
\newcommand{\chatgpt}{\model{ChatGPT}}
\newcommand{\gptfour}{\model{GPT-4}}

\newcommand{\llamatwo}{\model{LLaMA2}}

\NewDocumentCommand{\revanth}
{ mO{} }{\textcolor{blue}{\textsuperscript{\textit{Revanth}}\textsf{\textbf{\small[#1]}}}}
\title{Factcheck-Bench: Fine-Grained Evaluation Benchmark \\for Automatic Fact-checkers}

\author{Yuxia Wang$^{1}$, Revanth Gangi Reddy$^{2}$, Zain Muhammad Mujahid$^{1}$, Arnav Arora$^{3}$,  \\
        {\bf  Aleksandr Rubashevskii$^{1, 5}$, Jiahui Geng$^{1}$, Osama Mohammed Afzal$^{1}$, Liangming Pan$^{4}$}\\ 
        {\bf  Nadav Borenstein$^{3}$, Aditya Pillai$^{2}$, Isabelle Augenstein$^{3}$, Iryna Gurevych$^{1}$, Preslav Nakov$^{1}$}\\
  $^{1}$MBZUAI, Abu Dhabi, UAE \hspace{0.2em} $^{2}$University of Illinois at Urbana-Champaign \\$^{3}$University of Copenhagen, Denmark \hspace{0.2em} $^{4}$University of California, Santa Barbara 
  \\ $^{5}$Skolkovo Institute of Science and Technology, Moscow, Russia \\
  \texttt{\{yuxia.wang, preslav.nakov\}@mbzuai.ac.ae}
  }

\begin{document}
\maketitle
\begin{abstract}
The increased use of large language models (LLMs) across a variety of real-world applications calls for mechanisms to verify the factual accuracy of their outputs. 
In this work, we present a holistic end-to-end solution for annotating the factuality of LLM-generated responses, which encompasses a multi-stage annotation scheme designed to yield detailed labels concerning the verifiability and factual inconsistencies found in LLM outputs. 
We further construct an open-domain document-level factuality benchmark in three-level granularity: claim, sentence and document, aiming to facilitate the evaluation of automatic fact-checking systems.
Preliminary experiments show that \factool, \factscore and \perplexityai are struggling to identify false claims, with the best F1=0.63 by this annotation solution based on \gptfour.
Annotation tool, benchmark and code are available at 
\url{https://github.com/yuxiaw/Factcheck-GPT}.

\end{abstract}

\section{Introduction}
\label{sec:introduction}
Large language models (LLMs) have demonstrated impressive capabilities in terms of generating naturally-sounding answers over a broad range of human inquiries~\citep{OpenAI2023GPT4TR}, but
still frequently produce content that deviates from real-world facts~\citep{menick2022teach, bang2023multichatgpt, ali2023failurechatgpt, giuven2023llmsfailures,augenstein2023factuality}. 
This degrades the system performance and undermines its reliability, representing a significant bottleneck in their deployment especially for high-stake applications, e.g., clinical, legal, and financial settings~\citep{chuang2023dola}.

Before LLMs, most prior work investigate hallucinations of conditional text generation for specific tasks, such as abstract summarisation, dialogue generation, and machine translation~\citep{ji2023hallusurvey}.
They are either highly task-specific with gold standard references or focusing on short statements, in which automatic evaluation by rule-based matching or semantic similarity measurements with references is feasible.
However, in the case of free-form LLM generations over open domains, there is not a gold standard reference answer that can be employed to assess the factual correctness of model responses~\cite{wang2024factuality-survey}.
This makes the factual evaluation of open-domain LLM responses non-trivial, either depending on manual verification or automatic fact-checkers, e.g., \factscore and \factool~\cite{min2023factscore, chern2023factool}.
Human assessment demands substantial time and cost.

Despite surpassing humans in efficiency, the verification results of automatic fact-checkers are not necessarily accurate.
How to evaluate and improve the accuracy of automated fact-checkers is critical to produce dependable LLM factuality evaluations.

Many recent papers have built fact-checking systems and compared their evaluation results with human annotations. However, they only assess the final verification results, i.e., whether a claim or a document is true or false~\cite{chern2023factool, chen2023felm, dhuliawala2023chain}.
Regarding the system as a black box, without evaluations of intermediate results, makes it difficult to localise the specific component steps that eventually lead to the erroneous factual judgement. 
Desirable guidance for how to improve fact-checking pipelines is under-explored.
Moreover, prior studies use disparate evaluations, rendering them hard to compare and hampering future progress~\cite{wang2024factuality-survey}.

To address these issues, we introduce \textbf{Factcheck-Bench}, a new benchmark to evaluate automatic fact-checkers, with each example annotated with labels covering eight steps that would occur in the automatic detection and correction of factual errors, shown in~\figref{fig:pipeline}.
This enables evaluations of intermediate results of component modules in various automatic fact-checkers in a unified framework.
In sum, this work presents three major contributions: 
\begin{itemize}[noitemsep]
    \item We propose a holistic and fine-grained framework for annotating the factuality of the outputs of LLMs. It also serves as an end-to-end solution for detecting and correcting factual errors. 

    
    \item We construct a document-level claim-based fact-checking benchmark of LLMs, in terms of both detection and revision, facilitating the evaluation and analysis of automatic fact-checkers. In this process, we build an annotation tool to support efficient data collection. 
    
    \item We evaluate several popular checkers, such as \factool and \factscore using the annotated examples, and find large headroom for improvement in LLM fact-checking. We open-sourced the annotation tool, data, and code.
\end{itemize}

\section{Fine-grained Annotation Framework}
\label{sec:framework}

\begin{figure}[t!]
	\centering
	\includegraphics[scale=0.2]{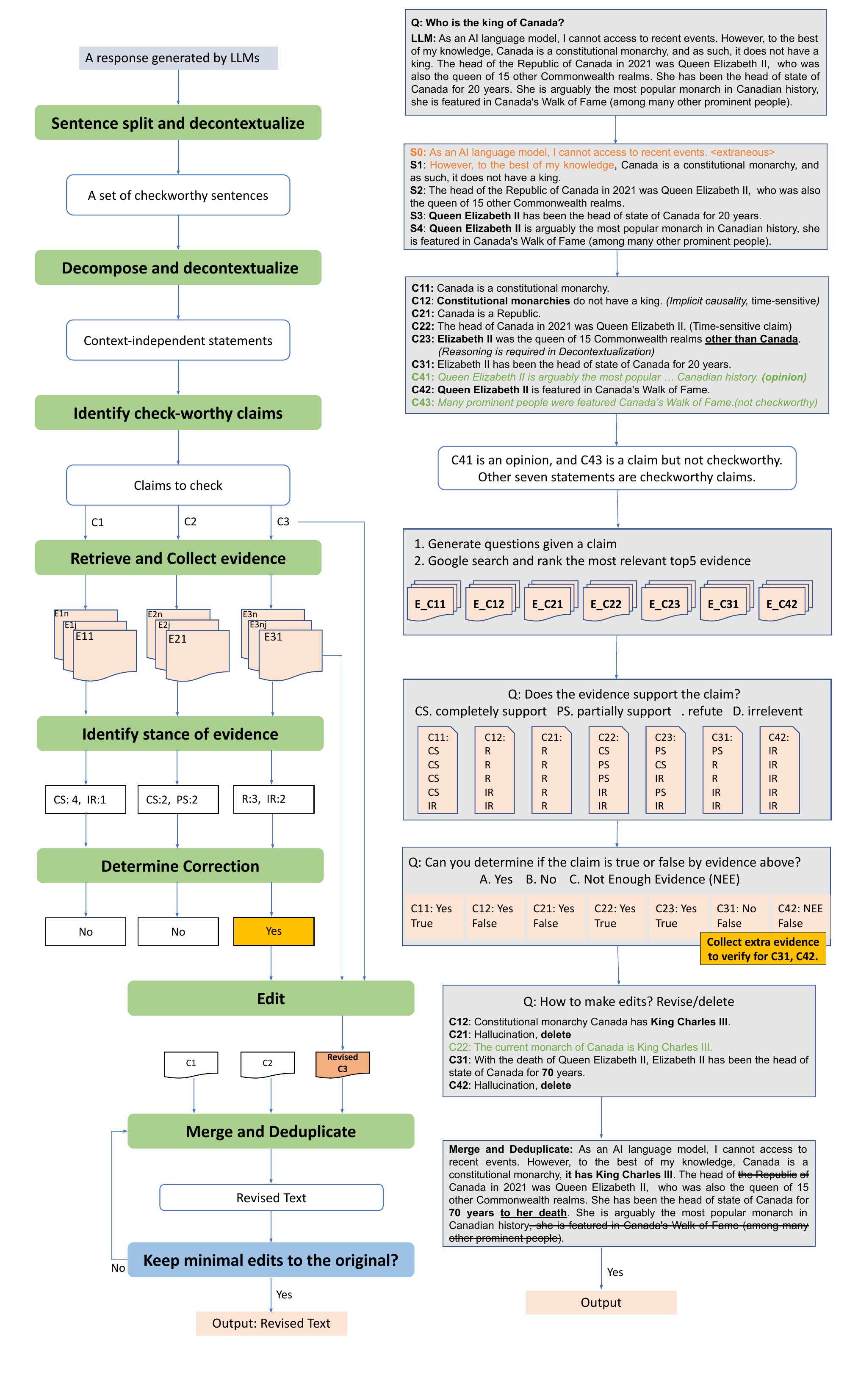}
	\caption{\textbf{Left:} Factuality annotation pipeline for LLMs outputs. \textbf{Right:} An example workflow.}
	\label{fig:pipeline}
\end{figure}

We frame the automated detection and correction of factual errors for outputs of LLMs into eight subtasks: (1) decomposition; (2) de\-contextualisation; (3) check\-worthiness identification; (4) evidence retrieval and collection; (5) stance detection; (6) correction determination; (7) claim correction and (8) final response revision. 
\figref{fig:pipeline} presents the overview of the whole procedure, coupled with an example flowing through each subtask.

\paragraph{(1) Decompose}
Given a response $R$ generated by a LLM, it is infeasible to fact-check the whole document at once, especially when it is long. The first step is to decompose $R$ into a set of context-independent atomic statements, with no information lost or distorted in this process.
Decomposed statements should be checkable independently without preceding and following context.\footnote{Statements are assumed to be checkable if relevant documents exist in publicly-available data sources.}

\paragraph{(2) Decontextualise} 
Sentences in a response might be context-dependent, with discourse and co\-reference relations existing between statements.
For example, it is invalid to check statement \textit{It does not have a king} before replacing ``It'' with ``Canada'' or ``Constitutional monarchy'' (see \figref{fig:pipeline}).   
In addition to co\-reference relation, for the sentence S2, it is not reasonable to check the claim \textit{Queen Elizabeth II is also the queen of 15 other Commonwealth realms.}
Instead, the claim should be re\-framed to \textit{Queen Elizabeth II is the queen of 16 Commonwealth realms (including Canada)} or \textit{Queen Elizabeth II was the queen of 15 Commonwealth realms other than Canada}. 

The concept of ``context-independent'' is straightforward, while the notion of ``atomic'' is subjective and ambiguous.
This poses challenges: how to determine the granularity of \textit{atomic claim}? when and where to break down a response?
For example, S1: \textit{Canada is a constitutional monarchy, and as such, it does not have a king}, can be fact-checked as one statement, or be decomposed into two claims: \textit{Canada is a constitutional monarchy} and \textit{Canada does not have a king}.
In our work, we first split a document into sentences, and then from sentence to claims, with each claim containing only one property or fact to verdict.

\paragraph{(3) Identify Check\-worthy Claims}
Not all statements in a response require fact-checking, such as subjective opinions and actual commonsense as obvious as \textit{sun rises from the east}.
Each statement in this framework will be identified whether it is check\-worthy or not.
However, check\-worthiness is subjective to determine.
\citet{hassan2015detectinh} defined check\-worthy claims as those for which the general public would be interested in knowing the truth.
In the context of fact-checking LLMs outputs, we assume users who ask LLMs questions are interested in knowing the truth of all factual claims in the corresponding answer.

We specifically classify a statement into four categories: factual claim, opinion, not a claim (e.g.\ questions, exclamations, imperatives), and others (e.g.\ \textit{As a language model, I cannot...}).
Afterwards, a set of checkworthy factual claims needs to be verified by retrieving and collecting evidence.

Note that for check\-worthiness, we not only take account of objective fact against subjective judgement, other aspects such as the role (importance) of the claim to the response is also a crucial criteria for its check\-worthiness.
For example, the sentence S1 needs more attention than the last sentence S4 in \figref{fig:pipeline}.
We label the importance level of both decomposed sentences and claims by labels: \textit{most important}, \textit{intermediate}, and \textit{less important}. 

\paragraph{(4) Retrieve and Collect Evidence}
Evidence can be retrieved by a search engine like Google, or deep retrieval from a closed document collection such as Wikipedia, or using the parametric knowledge of a LLM.
Search queries can be questions covering different aspects of the claim, entities in the claim, or even claim itself.
We used Google search considering the quality and coverage. 

\paragraph{(5) Identify Stance of Evidence}
With retrieved evidence for a claim, how to identify the stance of the evidence against the claim. 
RARR~\citep{gao2022attributed} achieved this by assessing whether answers depending on the evidence and the claim are same or not, given a query. 
If they are same, then the evidence supports the claim, otherwise refutes it.
Previous work also employs natural language inference (NLI) model to classify whether the claim can be entailed by evidence, or is controversial against evidence, or is irrelevant.

However, some evidence may neither refute nor fully support a claim.
This mainly results from the fact that it is always possible that the evidence supports part of the claim.
For example, for the claim \textit{Elon Musk is the founder, CEO, and chief engineer of SpaceX}, evidence \textit{Elon Musk is the CEO of SpaceX, Tesla, and Twitter} falls into this category.
The evidence supports the factual statement of \textit{Elon Musk is the CEO of SpaceX}, but it does not provide information regarding whether \textit{Elon Musk is the founder and chief engineer of SpaceX}.

Therefore, we incorporate \textit{partially support} in addition to \textit{support}, \textit{refute} and \textit{irrelevant}.
Concretely, \textit{support} means that the evidence entails the claim.
\textit{Partial support} refers to the scenario where part of the information presented in a claim appears in the evidence.
\textit{Refute} means that the evidence mentions the same event as the claim, but a clear opposite fact contrasting to a part or the whole facts presented in a claim. 
\textit{Irrelevant} refers to the situation that the evidence does not mention anything about the fact described in the claim, such that it neither supports nor refutes the claim.

Sometimes, it is ambiguous to distinguish between \textit{refute} and \textit{irrelevant}. We highlight that the evidence shows a clear opposite fact under \textit{refute} stance, while the evidence does not include relevant facts mentioned in the claim under \textit{irrelevant}.

\paragraph{(6) Determine Correction}
Given a claim, there will be more than one piece of related evidence.
Most of the time, they hold consistent stances except for irrelevance, but sometimes, some support, some partially support while some refute (see \figref{fig:conflicting-evidence-example}).
How to aggregate conflicting stances and further decide how to make corrections to the claim is an open question.
In practice, when evidence paragraphs conflict with each other, we will take the reliability of the evidence source into consideration and, meanwhile, retrieve extra information to judge which one is more dependable.

A label often used is \textit{not-enough-evidence} if there is insufficient information to make the veracity prediction, e.g., all retrieved evidence is irrelevant or intricate contradictory evidence~\cite{atanasova-etal-2022-fact}.
So we set three labels in terms of factuality: true, false, and not-enough-evidence.



\paragraph{(7) Edit Claims}
With the principle that revised claims should preserve the text’s original intent and style. without adding or changing unnecessary additional information, we include edit operations: delete the whole claim, replace X with Y, and delete X, where X and Y are meta information in a claim.

\paragraph{(8) Revise Response}
After revision, we merge statements in the original order, including non-check\-worthy statements, true claims, and revised claims. 
Finally, we delete reduplicative content if applicable, outputing a correct and fluent response.

\noindent \textbf{Discussion:} 
our annotation framework splits the fact-checking pipeline into eight steps, more fine-grained than existing systems. 
This intends to incorporate all subtasks and attributes relevant to automated fact-checkers, so that the comprehensive labels can cover evaluations of a wide array of unit modules within fact-checking systems. 
Practical implementation of fact-checkers can merge some steps, e.g., decomposition and decontextualisation into one, and evaluate the results of context-independent claims.

FELM~\citep{chen2023felm} annotated sentence-level \textit{true or false} labels without correction and showed that factual error detection performance tends to be improved when utilising claim-based segmentation methods compared with sentences.
Therefore, we annotate a claim-based document-level fact-checking dataset in \secref{sec:dataset}.



\section{Dataset Construction}
\label{sec:dataset}
We annotate a dataset serving for a benchmark evaluating the effectiveness of approaches for LLM fact-checking subtasks or the whole pipeline, and few-shot demonstration examples.

\subsection{Data Collection}
\textit{What kind of LLM generations are we most concerned about?}
In the context of detecting and correcting factual errors, we focus on generations in which the majority of statements are objective facts rather than subjective opinions whose veracity is not checkable.
Additionally, we are more interested in questions where LLMs are prone to hallucinate or produce factual errors in responses.
The whole annotation process is extremely time-consuming, about 15-30 minutes for an instance even if with the annotation tool to ease the procedure. 
This requests us to sample examples that highly satisfy two criteria --- \textbf{fact-intensive} and \textbf{factually-false}.

\paragraph{Sources}
We start from hallucinations posted by \chatgpt users on Twitter and further collect data by in-house brainstorming with preliminary verification, resulting in 45 examples.
We further employ data from dolly-15k, which is brainstormed by thousands of \textit{Databricks} employees with eight categories.\footnote{\chatgpt refers to GPT-3.5-turbo in this work.}
563 examples from closed QA and 528 from open QA are sampled by \chatgpt response length and the semantic similarity with gold answers, with 1,136 (question, response) pairs in total (see more in \appref{sec:datasource}).

\paragraph{Data Selection}
We select factually-false responses by estimating the percentage of incorrect claims in a response with four steps.

\textit{Sentence and claim split:}
given the whole response as the context and the first sentence (initialised by NLTK tokenizer), we instruct \chatgpt\ by three demonstration examples to guide it first breaking the input sentence into independent atomic claims, and then continue the decomposition of the next sentence until the end of the response (see the prompt in \appref{sec:atomicprompt}).

\textit{Evidence collection:} 
given a claim, we first prompt \chatgpt to generate search queries, and then the Google search engine is used to get relevant web pages. 
Retrieved documents are split into passages by sliding windows, and a re-ranker combining lexical and semantic similarity is used to identify the most relevant passages for the given query, in which Sentence-BERT~\citep{reimers-2019-sentence-bert} serves for semantic embeddings.
We aggregate evidence for all queries and select the top-5 evidences per atomic claim. 

\textit{\factscore calculation:}
\factscore is an automatic factuality metric, measuring 
the percentage of atomic claims supported by knowledge sources in a generation~\citep{min2023factscore}.
We use the gathered evidences as input, along with the claim, and an instruction-tuned LLM as the verifier to verdict.

\textit{Example selection:}
we keep all 45 pairs from the first source and dolly examples whose \factscore is less than 0.2, resulting in 33 closed QA pairs and 37 open questions, in total of 115 examples (see \factscore distribution in \figref{fig:factscoredist}).
We remove a similar question, and four questions where the LLM did not provide helpful answers due to its inherent disability to access real-time data, eventually annotating 110 examples. 



\subsection{Annotation} 
\label{sec:annotation}
Studies show that annotating a LLM factuality dataset is a highly challenging and time-consuming task~\citep{chen2023felm, li2023halueval}. 

\paragraph{Preliminary Trial Take-away}
manually annotating the whole process and typing results into a \textit{json} file exposes three major difficulties:
(1) retrieving supportive or contradictory evidence takes time and demands the annotator's strong skills in searching for relevant and filtering out unreliable information, especially for non-common knowledge (e.g.\ \textit{most popular bottled water brand in Israel});
(2) lengthy responses require good reading comprehension ability and patience; (3) certain domains such as genes and astronomy require domain knowledge, otherwise it is hard to search for valid evidence and determine whether it is true or false. 

Taking the factors mentioned above into consideration, we design and build an annotation tool to support the efficient construction of the LLM factuality benchmark.
Annotators can edit and assign labels based on intermediate outputs of automatic methods, click buttons instead of typing to copy-paste text, select, and download annotated results.\footnote{Without the annotation tool, on average, it takes $\sim$1.5 hours to annotate a 120-word response with about five bullet points, and more than 4 hours to annotate a 400-word response with ten bullet points, especially when the annotator is not familiar with details of an event (e.g., \textit{What are some details that are public about the 2021 Capitol Hill riots}). With the tool, it takes 15-30 minutes to label a 50-150 words document.}

\paragraph{Annotation Tool}
includes all subtasks and supports semi-auto annotation by incorporating the results of automatic methods, such as automatically-decomposed claims and automatically-retrieved evidence, to ease the annotation process and reduce the workload (see interfaces in \appref{sec:annotationinterfaces}).

We perform the whole annotation in three steps: 
(1) decomposition, decontextualisation, and check-worthiness detection;
(2) evidence stance identification and claim correction;
(3) claim merge, de\-duplication, and response revision.

Between steps (1) and (2), we incorporate an automatic evidence retrieval system to provide annotators with a set of the most relevant snippets of documents and URLs, as evidence for each check\-worthy atomic claim, generally five pieces.
They are selected and ranked by semantic relevance degree against the claim throughout a large number of documents, similar to the evidence collection in the data selection above.
Then, annotators determine the stance of each piece of evidence.
With evidence from the automatic system, if annotators still cannot determine the factuality of a claim, they are requested to collect relevant evidence manually.
This to some extent alleviates the system bias.


\paragraph{Quality Control}
To guarantee the annotation quality, instead of employing crowd-sourcing annotators, we perform an in-house labelling by ten annotators who are Master's and PhD students, postdocs, and professors and are familiar with fact-checking.

Two annotators as a group are responsible for 22 examples. 
For each step, annotators first independently finish individual annotations, and then consolidate their results with the group partner.
In consolidation, partners discuss their disagreements until reaching a consensus. For cases where it is hard to reach an agreement even with the participation of the third rater, we discard it.  
Three steps are rigorously conducted serially. Annotators start the second step only after they finish the consolidation of the first step.
Collecting evidence and judging stances is the most time- and patience-consuming step.
To ensure quality, we incorporate the third rater when consolidating the second-step annotations in case of unintentional mistakes.


\subsection{Data Analysis}
During annotation, we remove another 16 examples (see details in \appref{sec:dataanalysis}), resulting in 94 instances. Statistics are shown in \tabref{tab:statistics}.

\begin{table}[!t]
\centering
\resizebox{\columnwidth}{!}{
    \begin{tabular}{l c c c c c c}
        \toprule
        & document & sent & cw\_sent & claim & cw\_claim & evid \\
       \midrule
        size & 94 & 311 & 277 & 678 & 661 & 3,305 \\
        \bottomrule
    \end{tabular}
    }
    \caption{Statistics of the dataset. cw\_sent=checkworthy sentences, cw\_claim=checkworthy claims, evid=the total pieces of evidence, five for each cw\_claim.}
    \label{tab:statistics}
\end{table}
\paragraph{Statistics}
277 sentences contain factual statements among 311.
There are 678 atomic claims, where 661 claims are check\-worthy, 16 are opinions and one is \textit{not-a-claim}. 
For each checkworthy claim, five pieces of evidence are collected, resulting in 3,305 (claim, evidence, stance) triplets.

\textbf{How many examples are factually correct?}
61 examples contain factual errors, and 31 are factually correct, 2 without checkworthy claims.
Amongst, 53 examples contain false claims, and 19 examples contain claims in which annotators cannot verify the statement due to insufficient evidence despite the manual search.
Generally, one example contains 0-5 false claims. There are six examples with >5 incorrect claims.
16 sentences among 331 are deleted.
12 are total hallucinations, e.g., \textit{Trump was the second black president.} 
4 sentences are removed due to lacking enough evidence to support its factual correctness.

\tabref{tab:false-response-over-sources} shows that more incorrect responses appear in in-house collected questions, followed by dolly closed questions that require knowledge to obtain a unique correct answer. Fewer errors occur in dolly open questions, in which correct answers are not unique, e.g., \textit{How do you play an E major chord on a guitar?} It has diverse correct answers requiring more general knowledge.

\begin{table}[!t]
\centering
\resizebox{\columnwidth}{!}{
    \begin{tabular}{l c c c c}
        \toprule
        Source & In-house & Closed-QA & Open-QA & All\\
        \midrule
        Collected & 45 & 33 & 35 & 115\\
        Annotated & 39 & 30 & 25 & 94\\
        False & 38 & 16 & 8 & 61 \\
        \bottomrule
    \end{tabular}
    }
    \caption{False responses over three question sources.}
    \label{tab:false-response-over-sources}
\end{table}



\begin{figure}[t!]
	\centering
        \includegraphics[scale=0.285]{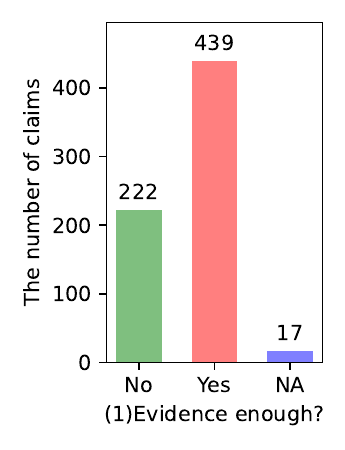}
        \includegraphics[scale=0.285]{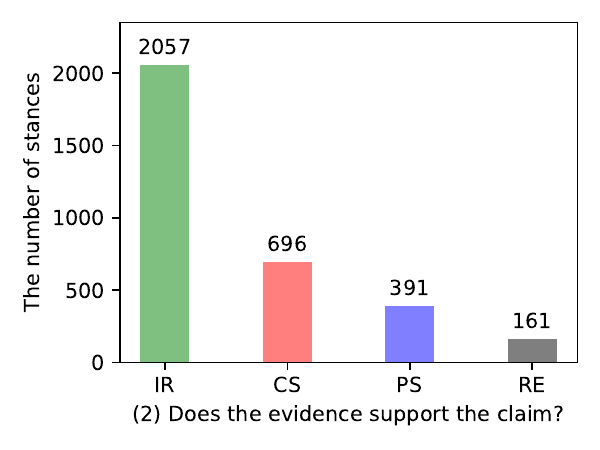}
        \includegraphics[scale=0.285]{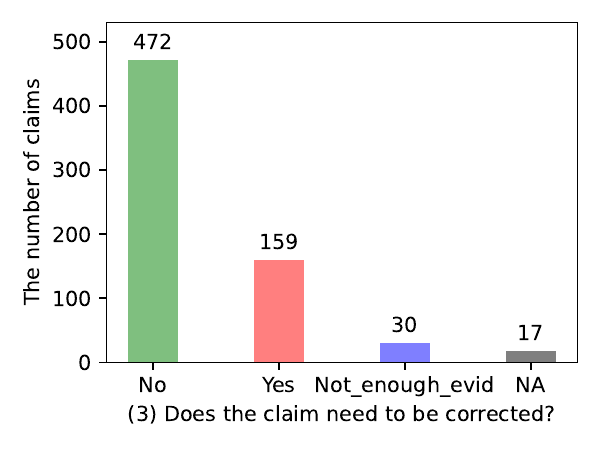}
	\caption{\textbf{Claim analysis:} (1) whether raters can determine the factuality of a claim depending on the automatically-collected evidence (\textit{Yes/No}); (2) does the evidence support the claim (\textit{CP}: completely support, \textit{PS}: partially support, \textit{RE}: refute, \textit{IR}: irrelevant); (3) does the claim need to be corrected. NA (17) refers to 16 opinion-claims + 1 \textit{not-a-claim}.}
	\label{fig:claim-dist}
\end{figure}
\paragraph{Claims} 
Of 678 claims, 419 and 227 are labelled as the most and intermediate important claims, and only 32 fall into \textit{not-important}, indicating that users concern with almost the whole response given their importance.
We analyse annotations of 661 checkworthy claims from two perspectives.

\textbf{Can raters determine the factuality of a claim depending on the automatically-collected evidence?} 
For 439 claims, annotators can determine \textit{true or false} with automatic evidence, while 222 claims (one-third) need further manual retrieval to make judgements.
Among the 222 claims, 125 true claims fall into domain knowledge and information that is less known by the external people given a country, region, company, or an individual.
The other half are either factually-incorrect claims (76) or undetermined claims without sufficient evidence despite manual retrieval (21), shown in \figref{fig:stance-dist}. 

This suggests the ineffectiveness of the automatic evidence retrieval methods on collecting rare knowledge and evidence conditioned on false premises (claims).
However, it may also reply that not all facts have been presented by textual descriptions directly. Some facts are unknown by the public, and some require connecting and reasoning knowledge from multiple sources, e.g., \textit{did Aristotle use a laptop?}~\citep{geva2021strategyqa}.

\textbf{How many claims need to be corrected?}
In \figref{fig:claim-dist}, about a quarter (159/661) of claims are factually incorrect and need to be corrected.
30 claims are undetermined due to inadequate related information and knowledge even with manual retrieval.
It is hard to obtain reliable related information about these cases by searching publicly-available sources. 
They involve expert-level knowledge (e.g., gene, water memory, black hole) and private details of an individual, organisation, or country (personal awards and preferences, revenue of a company), which are only known by a small group of people, such as domain experts or internal individuals who are familiar with the event.



\paragraph{Original vs. revised responses}
We quantify the difference between the original responses and the human-revised responses over the 61 false responses, showing that the normalised edit distance is 0.354, word overlap is 0.715, while semantically, BERTScore-F1 is 0.955 and cosine similarity based on SimCSE (Roberta-large) is 0.912.
This implies that the core content of LLM answers is mostly correct, but minor factual mistakes are easily made by LLMs in detail, leading to high semantic similarity but multiple lexical edits in small errors.

\paragraph{Summary} 
The dataset consists of 94 ChatGPT (prompt, response) pairs. Each sample has detailed labels concerning the verification: elements of decontextualised sentences, atomic claims, the importance degree of the sentence, claim to the response, five pieces of evidence for a claim, the relationship between a claim and evidence, factual label (\textit{true or false}) and revised version of claims, sentences, and the response.

\section{Unit Test for Fact-checkers}
In this section, we compare the results of automatic methods that are commonly used in current fact-checking systems (e.g., \rarr, \factscore, \factool) for subtasks with human annotations.
We first compare the automatic and human-annotated decomposition of atomic claim, and then evaluate five subtasks:
(1) identify whether the sentence contains a factual statement; 
(2) detect the check-worthiness of a claim by categories of \textit{factual}, \textit{opinion}, \textit{not a claim} and \textit{other};
(3) judge the stance of a given evidence against a claim, whether it \textit{supports}, \textit{partially supports}, \textit{refutes} or is \textit{irrelevant} to the claim;
(4) determine whether a claim is factually true or false, give a claim without ``gold evidence'', if false, revise it into a correct one;
(5) edit a list of originally-true or revised claims into a new response, given the original response, to correct the factual errors while preserving the linguistic features and style of the original. 

Other steps are excluded because they are either relatively easy for current techniques (e.g., splitting a document into sentences), or results of automatic approaches have been compared against human annotations in data analysis, such as the relevance or quality of the automatically-retrieved evidence. 

\subsection{Automatic vs. Manual Decomposition}
For 66/277 checkworthy sentences, the number of decomposed atomic claims is different between automatic breaking-down by \chatgpt and manual annotations.
Amongst, more claims decomposed by the automatic method than humans for 48 sentences, and fewer claims for 18 sentences.
This exhibits that human annotators add extra claims to only a small number of sentences.
In most cases, the automatic approach decomposes sentences into an equal number of claims or even more fine-grained than humans.

For the rest 211 sentences, human and \chatgpt decompose the sentence into the same number of claims, 521 claims are involved.
This enables pair-wise claim comparison between the human annotation and automatic method.
We calculate the lexical similarity and distance: normalised edit distance=0.11, n-gram distance=0.11, and word overlap=0.88 demonstrating high agreement between human annotation and \chatgpt decomposition.

\subsection{Checkworthiness}
We apply \chatgpt to identify if decomposed sentences and claims are verifiable objective facts or statements containing personal opinions.

\paragraph{Subtask 1 and 2}
We identify whether a sentence contains a factual statement by a binary label (\textit{yes} or \textit{no}) and whether a claim is checkworthy by four labels (\textit{factual claim}, \textit{opinion}, \textit{not-a-claim} and \textit{other}).
The accuracy for subtask 1 by majority guess (always checkworthy) will be 277/311=0.891 and the baseline for subtask 2: claim classification is 661/678 = 0.975.
They are superior to using the prompt based on \chatgpt: the accuracy is 0.814 and 0.932 respectively.
However, this is mainly attributed to the extremely-unbalanced data.
Practically, our aim is to make distinctions. It's critical to consider recall: \chatgpt is much better than the majority guess (see \tabref{tab:subtask1-2-results}).

The confusion matrix in \figref{fig:checkworthy-cm} shows that 46 checkworthy sentences are identified as non-checkworthy, accounting for 15\%. Factual claims could be recognised into any of the four labels, and real opinions tend to be identified as factual claims, even more than the opinion.

\begin{table}[!t]
\centering
\resizebox{\columnwidth}{!}{
    \begin{tabular}{l l c c c c}
        \toprule
        Task & Method & Acc & Prec & Recall & F1-macro \\
        \midrule
        1 & Always-checkworthy & 0.891 & 0.445 & 0.500 & 0.471\\
        1 & \chatgpt & 0.814 & 0.637 & 0.740 & 0.660 \\
        \midrule
        2 & Always-checkworthy & 0.975 & 0.325 & 0.333 & 0.329 \\
        2 & \chatgpt & 0.932 & 0.314 & 0.534 & 0.319 \\
        \bottomrule
    \end{tabular}
    }
    \caption{\textbf{Checkworthiness} detection by majority guess: Always-checkworthy vs. \chatgpt zero-shot prompt. \textit{average=}``macro'' is used in precision (Pred), recall and F1 calculation.} 
    \label{tab:subtask1-2-results}
\end{table}

\begin{table}[!t]
\centering
\resizebox{\columnwidth}{!}{
    \begin{tabular}{l c c c c}
        \toprule
        Method & Acc & Prec & Recall & F1-macro \\
        \midrule
        \multicolumn{5}{c}{\textbf{Four-label space}}\\
        Random guess & 0.255 & 0.258 & 0.264 & 0.215\\
        \llamatwo-zeroshot & 0.202 & 0.324 & 0.280 & 0.155 \\
        \chatgpt-zeroshot & 0.365 & 0.402 & 0.439 & 0.332 \\
        \midrule
        \multicolumn{5}{c}{\textbf{Three-label space}} \\
        \chatgpt-zeroshot & 0.567 & 0.506 & 0.588 & 0.483 \\
        \llamatwo-zeroshot & 0.401 & 0.407 & 0.384 & 0.299 \\
        RoBERTa-large-mnli & \textbf{0.607} & \textbf{0.536} & \textbf{0.609} & \textbf{0.512} \\
        \bottomrule
    \end{tabular}
    }
    \caption{\textbf{Stance} detection by \chatgpt and \llamatwo zero-shot prompt. Three-label space merges complete and partial support into one.} 
    \label{tab:subtask3-results}
\end{table}

\subsection{Verification}
\paragraph{Subtask 3} classifies whether the evidence fully supports, partly supports, refutes, or is irrelevant to the claim, given a \textit{(claim, evidence)} pair. 
We use zero-shot prompting based on \chatgpt and \llamatwo (7B), and find that \llamatwo barely predicts \textit{partial support} and always misclassifies as \textit{irrelevant}, so we merge \textit{complete support} and \textit{partial support} into a single label \textit{support}.
As results shown in \tabref{tab:subtask3-results}, three labels are easier for models to predict with higher accuracy, but its absolute F1-score is still less than 0.5, revealing the challenges to distinguish the relationship between claim and evidence by LLM in-context learning, especially on the label of \textit{refute}.
Both \llamatwo and \chatgpt show around-0.1 F1 (see \tabref{tab:subtask3-results-label-specific}).
We further use a fine-tuned NLI model (\textit{RoBERTa-large-mnli}) to predict the stance, where entailment, contradiction, and neutral correspond to labels of support, refute, and irrelevant respectively.
It performs better than zero-shot \chatgpt, mainly being superior to predicting the label of \textit{support}. 

\textbf{Subtask 4} determines whether the claim is true or false by leveraging the evidences retrieved from external knowledge sources.
We evaluate the verification methods used in \factscore~\citep{min2023factscore} and \factool~\citep{chern2023factool}, with varying evidence sources: Wikipedia (September 2023 dump) and web articles searched by Google. 
Commercial verifier \perplexityai and the verifier implemented with Google search + \gptfour based on the solution in this work (Factcheck-GPT) are also evaluated.

\tabref{tab:subtask_4} shows that false claims tend to be identified less accurately than true claims across all approaches, implying that it is more difficult to detect factual errors than the correct statements.
Factcheck-GPT performs the best on false claims with F1=0.63, and then Perplexity.ai by 0.53, followed by Instruction-LLaMA with web articles as evidence (F1=0.47/0.84), and verifying using GPT-3.5-Turbo exhibits slight declines. 
This reveals that current mainstreaming SOTA fact-checkers still have large room to improve on verification, particularly on false claims.  
Performance using Wikipedia as the source is inferior to using web articles, this is largely limited by the knowledge coverage of Wikipedia, esp. on open-domain benchmarks.

\begin{table}[t!]
\scriptsize
    \centering
    \resizebox{\columnwidth}{!}{
    \begin{tabular}{l|c|ccc|ccc}
    \toprule
    \multicolumn{1}{c|}{\multirow{2}{*}{\textbf{Verifier}}} & \multicolumn{1}{c|}{\multirow{2}{*}{\textbf{Source}}} & \multicolumn{3}{c|}{\textbf{Label = True}} & \multicolumn{3}{c}{\textbf{Label = False}} \\
    & & Prec & Recall & F1 & Prec & Recall & F1 \\
    \midrule
    Random & NA & 0.79 & 0.43 & 0.56 & 0.18 & 0.52 & 0.27\\
    Always True & NA & 0.81 & 1.00 & 0.88 & 0.00 & 0.00 & 0.00 \\
    Always False & NA & 0.00 & 0.00 & 0.00 & 0.19 & 1.00 & 0.33 \\
    \midrule
    Inst-LLAMA & Wiki & 0.87 & 0.74 & 0.80 & 0.34 & 0.56 & 0.42\\
    Inst-LLAMA & Web & 0.88 & 0.80 & 0.84 & 0.40 & 0.56 & 0.47\\
    GPT-3.5-Turbo & Wiki & 0.87 & 0.67 & 0.76 & 0.31 & 0.60 & 0.41\\
    GPT-3.5-Turbo & Web & 0.89 & 0.74 & 0.81 & 0.37 & 0.62 & 0.46\\
    \midrule
    Perplexity.ai & Web & \textbf{0.93} & \textbf{0.73} & \textbf{0.83} & 0.40 & 0.76 & 0.53 \\
    Factcheck-GPT & Web & 0.90 & 0.71 & 0.79 & \textbf{0.52} & \textbf{0.80} & \textbf{0.63} \\
    \bottomrule
    \end{tabular}}
    \caption{\textbf{Verification results} on our benchmark: judge whether a claim is factually true or false with external knowledge (Wikipedia or Web articles) as evidence.}
    \label{tab:subtask_4}
    \vspace{-1em}
\end{table}

\subsection{Revision}
\paragraph{Subtask 5} Given the original false response, a list of revised true claims, \chatgpt and \gptfour are prompted to revise the responses with/without the question, resulting in four revised responses.

Which revised response is better?
We evaluate by human and the intrinsic metrics. BERTScore measures semantic preservation between gold reference answers and the edit-distance measures style preservation between original responses. 

In human evaluation, we use the criteria: whether the revised response 
(1) contain factual errors?
(2) keep the style feature of the original response as much as possible?
(3) is it natural, coherent, and smooth as an answer?
Criteria (1) is the most important, followed by (2) and (3).
For instance, only \textit{A} and \textit{B} are factually correct, while \textit{A} preserves more of the original response, thus \textit{A} is better. If some responses are totally the same, raters can choose more than one. We collect 66 preference labels for 61 examples. 

In case of personal preference bias from one or two raters, six raters are invited to choose their preferred response and provide a brief reason. 
We also shuffled four revisions and show by ``revision\_x'' (x=0,1,2,3), masking the real setting name to avoid possible inherent biases.

In \tabref{tab:subtask5-results}, intrinsic metric results show that responses revised by \chatgpt (GPT-3.5-turbo) are better than \gptfour, which is against our experience and observation (see examples in \appref{sec:subtask5-prompt}).
Human assessment exhibits that 43 \gptfour responses are preferred by raters and 23 from \chatgpt. 
Human is more satisfied with revisions without questions 38 vs. 28. 
This somewhat reflects the ineffectiveness of intrinsic evaluation metrics. 

\begin{table}[!t]
\centering
\resizebox{\columnwidth}{!}{
    \begin{tabular}{ll cccc c}
        \toprule
        Prompt & model & Edit-dis$\downarrow$ & WO$\uparrow$ & BS-F1$\uparrow$ & STS$\uparrow$ & Human \\
        \midrule
        no-ques & \chatgpt & \textbf{0.207} & \textbf{0.864} & 0.953 & 0.937 & 10\\
        no-ques & \gptfour & 0.275 & 0.789 & 0.954 & 0.931 & 28 \\
        with-ques & \chatgpt & 0.222 & 0.853 & \textbf{0.956} & \textbf{0.941} & 13\\
        with-ques & \gptfour & 0.286 & 0.776 & 0.953 & 0.935 & 15 \\
        \bottomrule
    \end{tabular}
    }
    \caption{\textbf{Revision evaluation} by intrinsic metrics and human (how many responses are preferred). Edit distance (\textbf{Edit-dis}) and word overlap (\textbf{WO}) between revised and the original responses. BERTScore (\textbf{BS-F1}) and semantic textual similarity (\textbf{STS}) based on SimCSE between the revised responses and human annotations.}
    \label{tab:subtask5-results}
\end{table}


\paragraph{Take-Away}
\chatgpt shows promising results in atomic-claim decomposition, but low F1-score in checkworthiness detection.
Also, verification remains challenging, especially when identifying false claims, even if it involves harnessing external knowledge.
\gptfour can generate sounding revised responses based on true statements. It's still an open-question in terms of how to evaluate the quality of revised responses by intrinsic metrics.
\section{Conclusion}
We propose a fine-grained annotation framework and construct a benchmark to evaluate automatic fact-checkers.
The benchmark collects 678 open-domain claims of LLMs, involving annotations of eight subtasks for detecting and correcting factual errors in long documents. 
Human annotations show that LLMs are prone to make factual errors in expert-level knowledge and exclusive details known by a small group of people. 
Experiments show that current verifiers are struggling to identify open-domain false claims with the best F1=0.63 even if using external knowledge. 
Additionally, intrinsic metrics based on edit distance and semantic similarity are ineffective in evaluating the edited responses against true evidence and the original response, misaligning with human preferences. It is worth more exploration in the future work.

\section*{Limitations}
Three major limitations are identified in this work:

\paragraph{Small-scale dataset} It consists of only 94 (question, response) pairs, we plan to scale up the dataset in English, Chinese, and Arabic in future work.
It is worthwhile to note that our dataset contains fine-grained annotations of high quality for eight subtasks.
Moreover, due to high cost (i.e., over 30-50 USD on average to evaluate 100 LLM responses depending on its length), developers generally evaluate on less than 100 examples in the development iterations, sometimes even 10 examples to save costs. Practically, our dataset is enough to be used as a benchmark for the preliminary evaluation of automatic fact-checking systems.

\paragraph{Inter-claim dependencies}
This reflects at three challenges.
First, current approaches including our solution are unable to check the overall logical correctness of a procedure, such as how to cook, and whether some steps are out of order.
Second, if the first claim is invalidated, maybe the entire text needs to be deleted.
Third, it is hard to decontextualize implicit claims, e.g., ``other 15 realms'', which means there are 16 realms.

\paragraph{Quality of evidence}
More than half of automatically retrieved evidences are irrelevant.
Improving the relevance of retrieved evidence is critical to the accuracy of fact-checking.

\section*{Ethics and Broader Impact}
We identify two major risks of the framework and benchmark:

\paragraph{Biases:}
The automatic atomic-claim decomposition and evidence retrieval systems incorporated in the fact-checking annotations may introduce biases, which can affect the annotation results.

Besides, the dataset does not cover all types of claims. Limited scope and coverage may be more effective in certain domains, possibly leading to inaccurate or unfair assessments in certain domains for automatic fact-checkers.
The responses generated by LLMs might also inherit some biases present in the involved LLMs. 

\paragraph{The cost of making an error:} 
The goal of fact-checking is to improve the reliability of the LLM outputs, 
If post-hoc fact-checking methods under this framework always make errors, practitioners may lose faith in the accuracy of the fact-checking results, which can affect efforts to maintain public trust in fact-checking systems.

\paragraph{Broader impact:} The proposed framework is not limited to checking the output of LLMs; it is applicable to checking any kind of document, including human-written.

\bibliography{ref}
\bibliographystyle{acl_natbib}

\clearpage
\onecolumn
\section*{Appendix}
\appendix
\section{Fact-checking Background}
\label{sec:background}
\subsection{What is Fact-checking?}
Fact-checking is the task of assessing whether claims made in writing are manipulated or true.
This is typically broken down into the stages of claim detection, evidence retrieval, verdict prediction, and optionally justification prediction \citet{guo-etal-2022-survey,Augenstein2021Doctoral}.

Claim detection is to identify claims that require verification, which commonly relies on the concept of check-worthiness. In the context of human-written documents, checkworthy claims are regarded as those for which the general public would be interested in knowing the truth~\citep{hassan2015detectinh,wright-augenstein-2020-claim}. 
However, this may not be adaptable to machine-generated texts.
Plausible hallucinations of LLMs make it difficult for general individuals to distinguish whether it is true or false, thus their outputs become less trustworthy than the statements made by humans. Current methods tend to check all factual claims of LLM generations~\citep{chern2023factool}.

Evidence retrieval aims to find sources supporting or refuting the claim.
Claim verification is expected to assess the veracity of the claim and produce justification based on the retrieved evidence. That is, claims are assigned truthfulness labels, and explanations for verdicts are produced.
A basic form of justification is to highlight the pieces of evidence used to reach a verdict~\citep{guo-etal-2022-survey}.

\begin{table*}[ht!]
    \centering
    \resizebox{\textwidth}{!}{
    \begin{tabular}{lccllcp{4cm}p{8cm}}
    \toprule
    Method & \textbf{D} & \textbf{R} & Granularity & Knowledge source & Datasets & Task & How\_collect \\
    \midrule
    Factcheck-GPT & \cmark & \cmark & claim & Google search & \cmark & Instruction & prompt \chatgpt and human annotation \\
    \factool~\citep{chern2023factool} & \cmark & \cmark & article metadata & Google scholar & \cmark & Generate literature review & prompt \chatgpt \\
    \factool~\citep{chern2023factool} & \cmark & \cmark & claim (gold) & Parsed Google search & RoSE/FactPrompts & Summarisation-eval/QA & human annotation: RoSE~\citep{liu-etal-2023-revisiting} \\
    \rarr~\citep{gao2022attributed} & \cmark & \cmark & document & Bing search & NQ,StrategyQA,QReCC & QA & human annotation\\
    \cove~\citep{dhuliawala2023chain} & \cmark & \cmark & document & parametric knowledge & CoVe corpus & QA, instruction  & human annotation \\
    
    \midrule
    FELM~\citep{chen2023felm} & \cmark & \xmark & segment & Google search & \cmark & Instruction & prompt \chatgpt and human annotate factuality \\
    Self-contradictory~\citep{niels2023selfcontradictory} & \cmark & \xmark & sentence & parametric knowledge & \cmark & Instruction & prompt \chatgpt,\gptfour for contradictory sentence \\
    SelfCheckGPT~\citep{manakul2023selfcheckgpt} & \cmark & \xmark & sentence & parametric knowledge & \cmark & Generate Wikibio passage & prompt GPT3 and human annotate 3 factual labels   \\
    \factor~\citep{muhlgay2023factor} & \cmark & \xmark & sentence & parametric knowledge & \cmark & Multichoice QA & prompt \textit{davinci-003} for non-factual completions \\
    HaluEval~\citep{li2023halueval} & \cmark & \xmark & document & parametric knowledge & \cmark & QA, summarise, dialogue & prompt \chatgpt to generate hallucinated answers \\
    HaluEval~\citep{li2023halueval} & \cmark & \xmark & document & parametric knowledge & \cmark & Instruction & prompt \chatgpt, human annotate false segments \\
    \factscore~\cite{min2023factscore} & \cmark & \xmark & claim & Wiki Bio Generation & \cmark & Instruction & prompt \chatgpt to generate biography\\ 
    FRESHQA~\citep{vu2023freshllms} & \cmark & \xmark & facts in answer & parametric knowledge & \cmark & QA & collect questions with time-changing answers\\
    Snowball~\citep{zhang2023snowball} & \cmark & \xmark & Yes/No answer & parametric knowledge & \cmark & QA & human annotation \\ 
    SelfAware~\citep{yin-etal-2023-large} & \cmark & \xmark & document & reference generations & \cmark & QA & collect unanswerable questions and prompt \chatgpt \\ 
    \bottomrule
    \end{tabular}}
    \caption{\textbf{Methods and benchmarks for hallucination Detection (D) and Revision (R)}. \factool: article metadata is a tuple (paper title, year, authors). CoVe=Chain-of-Verification, CoVe corpus includes four existing datasets: Wikidata, Wiki-category, MultiSpanQA, and biographic. 3 labels in SelfCheckGPT: major/minor inaccurate and accurate. Unanswerable questions: the model should express uncertainty instead of delivering conclusive responses. FRESHQA collect four types of questions: false premise, answers never change, change slowly and fast over time.}
    \label{tab:relatedwork}
\end{table*}

\subsection{Conventional Fact-checking}
Previous works either focus on hallucinations in task-specific generations with references (to detect whether the generated output contradicts the source content), such as abstractive summarization~\citep{maynez-etal-2020-faithfulness}, machine translation~\citep{raunak-etal-2021-curious} and data-to-text generation~\citep{liu2021tabletotext},
or concentrate on specific topics e.g.\ Covid-19~\citep{augenstein-etal-2019-multifc}, politics~\citep{barrera2020facts}, climate~\citep{thomas2020climatefever}, and specific domains such as journalism, news, social media (e.g.\ Twitter~\citep{nicolas2022twitter}) and Wikipedia (FEVER: \citet{thorne-etal-2018-fever}). In contrast, we set target for text generation tasks without references such as generative question answering and dialogue systems in open domain and open topic, where the source is the world knowledge.

Moreover, most early studies only perform one or two subtasks in the factual error detection and correction, instead of the whole process.
For example, many models estimate a label indicating whether the statement is supported or refuted by the evidence, given a (statement, evidence) pair as input~\citep{thorne-etal-2018-fever, nie2019combine, augenstein-etal-2019-multifc, wadden-etal-2020-fact}. 
To adapt to situations where relevant evidence for a statement is not readily available, some works explored how to automatically retrieve evidence that may help support or refute a statement~\citep{fan-etal-2020-generating, preslav2021automated, gao2022attributed}.

More recent work has also explored how to correct claims based on retrieved evidence~\citep{thorne-vlachos-2021-evidence, schuster-etal-2021-get, iv-etal-2022-fruit} and how to generate justification/explanation for verdicts on claims~\citep{atanasova-etal-2020-generating-fact}. 
However, most factual correction used human-authored edits from FEVER~\citep{thorne-etal-2018-fever} as both their training and automatic evaluation data. FEVER’s claims were extracted from Wikipedia. This limits the generalisability of these fact-checking models over generations of LLMs across various tasks and diverse domains.

Our goal is to establish a holistic framework, evaluating systems that automatically detect and correct factual errors end to end for open-domain factual knowledge hallucinations.


\subsection{LLM Fact-checking}
The phenomenon that LLMs produce outputs that are seemingly plausible while deviating from the user input, previously generated context, or factual knowledge, is commonly referred to as hallucination~\citep{zhang2023siren}.
Based on the timing of the LLM life cycle, LLM hallucinations can be addressed during pretraining by automatically selecting reliable data or filtering out noisy data to mitigate hallucinations, in supervised fine-tuning by curating a small volume of high-quality training data, in reinforcement learning from human feedback (RLHF), and during inference by decoding strategies~\citep{zhang2023siren}.
We focus on approaches applied after inference. 


\paragraph{Methods}
For post-processing approaches to alleviating LLM hallucinations, recent studies can be roughly classified into two categories depending on whether they rectify errors: (1) detecting and correcting factual errors for free-form text; and (2) only detecting whether a text contain hallucinations (\textit{Yes} or \textit{No}).
Both of them resort to either external knowledge or parametric knowledge to identify and rectify factually-incorrect statements~\citep{gao2022attributed, chern2023factool, manakul2023selfcheckgpt, dhuliawala2023chain}.
We used external knowledge retrieved from Google.

Our work puts efforts into facilitating the first category.
Though Self-contradictory~\citep{niels2023selfcontradictory} involves revision, they aim to remove the conflicting information between the original sentence and the synthetically-generated contradictory sentence, instead of correcting factual errors in the original sentences.
We classify it into the second category: detection only.
\rarr~\citep{gao2022attributed}, \factool~\citep{chern2023factool} and \cove~\citep{dhuliawala2023chain} are three most relevant work to ours.

Given a LLM response, \rarr and \cove first generate a series of questions covering different aspects of the response, which serve as queries in the evidence retrieval, and then edit the whole response to correct factual errors.
Such coarse granularity verification may miss out incorrect statements, particularly over long documents, and also makes it difficult to spot false spans precisely, thus disabling fine-grained (e.g., correct only a false number in a statement) and flexible edits (e.g., delete a completely-hallucinated sentence).
Additionally, revising the whole document tends to result in poor preservation of the original input (e.g., style, vocabulary, and structure), introducing irrelevant descriptions and even new hallucinations.
Claim-level editing empowers precise correction and good preservation.

\factool performs fact-checking over claims.
However, gold claims are required as input for the system.
That is, users must first decompose an output from a LLM into a list of checkable atomic claims by themselves before using \factool to check, which complicates the fact-checking process.
Moreover, it is expensive to use \factool to check a piece of text, since the whole checking process calls APIs including OpenAI (\$0.06/1K tokens), Serper (\$1.00/1k queries), and Scraper.\footnote{\url{https://www.scraperapi.com/pricing/}}
This also challenges the evaluation where online API is not allowed to call with the consideration of internal data protection.

We attempt to alleviate these issues in our framework.
We decompose the fact-checking task into eight subtasks. 
The design of decomposing and decontextualising a long document into independent sentences and then into atomic claims allows inputs of any granularity: document, sentence, or claim.
The pipeline equipped with check-worthiness selection also naturally endows the flexibly-customised verification, such as skipping subjective statement, commonsense and the knowledge is well-known by the individual.

\begin{table}[ht!]
    \centering
    \resizebox{0.7\columnwidth}{!}{
    \begin{tabular}{l c c c c r}
    \toprule
    Dataset & Granularity & Factual label & Revision & Length & Size \\
    \midrule
    HaluEval & document & \cmark & \xmark & 82.0 & 4,507\\
    FELM-WK & segment & \cmark & \xmark & 51.1 & 184 \\
    FactPrompts & claim & \cmark & \xmark & 41.8 & 50 \\
    \midrule
    Factcheck-GPT & claim & \cmark & \cmark & 73.1 & 94 \\
    \bottomrule
    \end{tabular}}
    \caption{Statistics of world-knowledge factuality evaluation benchmarks. Length=the average number of words of LLM responses.}
    \label{tab:wkqa}
\end{table}

\paragraph{Datasets}
From the perspective of the evaluated benchmarks, as shown in \tabref{tab:relatedwork}, studies of the first category generally evaluate their methods on existing QA datasets, or revise hallucinations in a specific topic such as literature review and biographic generations~\citep{chern2023factool, dhuliawala2023chain}.
These topics may not be frequently requested by general users in real-world scenarios.

Studies of the second category contribute a spectrum of benchmarks to detect diverse hallucinations, such as synthetically-generated contradictory sentences~\citep{niels2023selfcontradictory}, deliberately-generated hallucinated answers~\citep{li2023halueval} and non-factual completions given a prefix context~\citep{muhlgay2023factor}.
\citet{manakul2023selfcheckgpt} manually annotate factual labels (major/minor inaccurate and accurate) given a sentence in the generated Wikibio passage.

Interestingly, \citet{yin-etal-2023-large} collected 1,032 unanswerable questions from five diverse categories \textit{no scientific consensus}, \textit{imagination}, \textit{completely subjective}, \textit{too many variables}, \textit{philosophical}, and their 2,337 answerable counterparts.
Unanswerable questions refer to questions where the model should express uncertainty instead of delivering conclusive responses.
\citet{zhang2023snowball} collected three datasets, with 500 questions (all \textit{No} or all \textit{Yes} answers) for each. One focuses on one type of question, including whether a number is a prime, senator search (whether a US city has a specific university), and whether there is a flight from one city to another given a graph connection.

However, these datasets are either only applicable in detection, or originate from a single task like biography writing~\citep{min2023factscore}, without accounting for variations across different generations.
HaluEval's annotation over Alpaca 5K responses of various instructions, which is one of the most similar work to ours.
They ask human annotators to label whether the response contains hallucinated information (\textit{Yes} or \textit{No}) and list the corresponding spans if there exist errors~\citep{li2023halueval}.\footnote{The hallucination is considered from the following three aspects: unverifiable, non-factual, and irrelevant.}
FELM with 184 world-knowledge questions is labelled in the granularity of segments, while ours are over fine-grained claims to locate factual errors more precisely.
Moreover, our annotations not only include factual labels of each claim, but the revised text and labels of all involved subtasks as well, e.g., decomposition of a sentence into a list of independent claims, check-worthiness of a sentence/claim, evidence stance and so on.

\factool evaluate over a knowledge-based QA dataset FactPrompts consisting of 50 (prompt, response) pairs.
It is annotated by authors over atomic claims and their factual labels (true/false), but the responses tend to be short, instead of long documents (see \tabref{tab:wkqa}).
Overall, our dataset offers both factual labels and the revised text in three-level granularity --- atomic claims, decontextualised sentences, and responses, for LLM answers, with an emphasis on long documents.


FELM~\citep{chen2023felm} is the most relevant concurrent work with ours, but only annotated sentence-level \textit{true or false} labels (no correction).
We construct a new dataset that collects (question, ChatGPT response) pairs in real conversations. 
Annotators identify and edit factual errors for each atomic claim decomposed and decontextualised from the original long-form responses.
This is expected to serve as a benchmark to evaluate the performance of fact-checkers.

\clearpage
\section{Dataset}
\subsection{Sources}
\label{sec:datasource}


\paragraph{Twitter posts and in-house brainstorming:} 
We first collect (question, response) pairs from ChatGPT/GPT-4 failures found on social media, in Web articles, and in related papers.\footnote{\url{https://github.com/giuven95/chatgpt-failures}}
The query should satisfy the criteria that the corresponding response must have factual errors, rather than failures regarding reasoning, math, coding, bias, and so on; (query, response) also should be independent of a dialog. This results in 23 examples. 
We additionally brainstorm a spectrum of questions depending on individual usage experience of ChatGPT and then select 22 questions whose responses contain factually-false content by manually verifying suspicious facts.

\paragraph{Dolly-15k} 
It consists of 15,011 examples, with eight categories ranging from closed, open, and general QA, to creative writing, brainstorming, information extraction, summarisation and classification.\footnote{Its use is subject to the \textit{CC BY-SA 3.0} license.}
Since we pay attention to open-domain generations and responses with more factual statements, closed and open-question answering pairs are chosen to be the database.

We first generate ChatGPT responses for 1,773 closed QA pairs without using context information (a paragraph extracted from Wikipedia relevant to the question), and 3,700 open QA pairs.
After filtering questions that cannot be answered without context as well as questions ChatGPT does not answer, we further filtered responses with fewer than 200 characters.
Taking human answers as the gold reference, we assume that if machine generations are semantically far from human answers, they may contain false information.   
So we keep the examples where the cosine similarity <= 0.5 between human answer and machine response based on SimCSE sentence embedding.
Finally, we select 563 examples from closed QA and 528 from open QA, thus 1,136 (question, response) pairs in total with 45 from the first source.

\subsection{Data Selection} 
\label{sec:dataselection}
The whole annotation process is extremely time-consuming, about 15-30 minutes, for an instance, even if with intermediate results from automatic methods to ease the procedure.
This requests us to cherry-pick examples that highly satisfy two criteria --- fact-intensive and factually-false.
Therefore, we leverage \factscore to filter cases with the following four steps.

\paragraph{Sentence split and atomic claims breaking-down}
We first split a document into sentences using the NLTK tokenizer.
The most straightforward way is to prompt \chatgpt\ to split a sentence into claims given the response as context. However, processing sentences one by one consumes both time and API tokens.

Therefore, given the whole response as the context and the first sentence of the response, we ask \chatgpt\ to break the input sentence into independent atomic claims, and also continue the decomposition of the next sentence of the response (see the prompt in \secref{sec:atomicprompt}).
Specifically, \chatgpt\ is given three demonstration examples, so that it can follow the instruction to first break down the input sentence into atomic claims, and then sequentially find the next sentence and make the splits.
Over 90\% examples follow the instruction, breaking down the whole response.
105 out of 1,136 examples only decompose the first sentence, on which we process sentence by sentence based on the NLTK sentence splits.

Another reason why we ask \chatgpt\ to re-split the response into single sentences is that we observed that some sentences are incorrectly split into smaller units by NLTK, such as decomposing a paper reference into a set of metadata, while \chatgpt\ can remain the citation reference as a whole.\footnote{In our dataset, we prioritise sentence splits by \chatgpt, using NLTK results for unsuccessfully-parsed instances. The prompt is initialised with the first sentence split by NLTK.}
A weakness of \chatgpt\ outputs compared with traditional models is that it is sometimes non-trivial to parse the results from the text-free responses when \chatgpt\ does not follow the output format as the instruction.
In such cases, we have to process examples specifically.
\\
\textbf{\textit{Discussion:}}
One may argue that why not directly decompose the whole response into atomic claims, but through single sentences and then to atomic claims? There are two reasons.
\begin{itemize}[noitemsep]
    \item \textit{Avoid distortion:} atomic claims decomposed and decontextualized from a whole response by models such as \chatgpt\ tend to either lose or hallucinate information compared to the original response, while the quality of atomic claims of a single sentence is much better;
    \item \textit{Improve annotation quality:} sentences as the intermediate state, it is easier for annotators to go through 1-5 atomic claims for a sentence as one annotation unit, instead of >5 claims for a whole response (particularly long documents), which helps annotators to pay attention and improve the annotation quality.
\end{itemize}

\begin{figure*}[t!]
	\centering
	\includegraphics[scale=0.42]{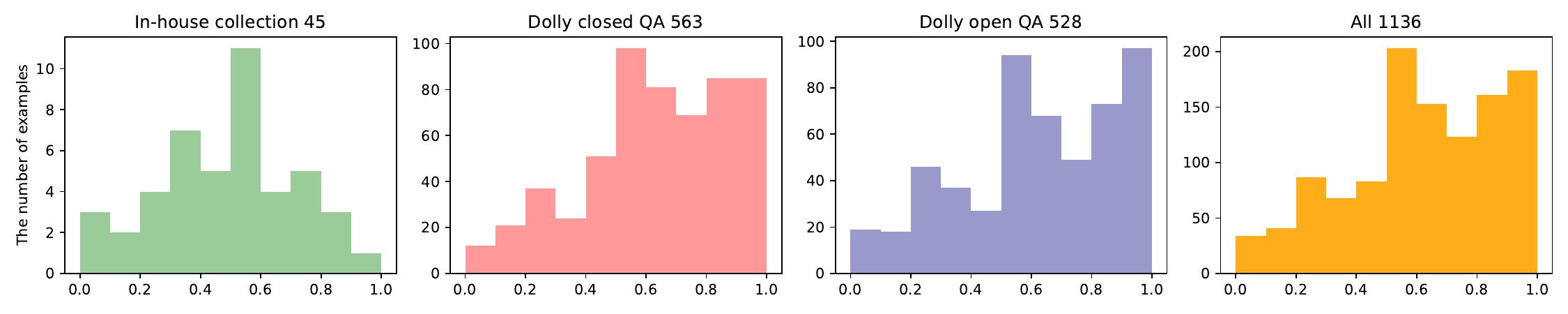}
	\caption{\factscore distribution for three component sources and their combination.}
	\label{fig:factscoredist}
\end{figure*}
\paragraph{Evidence collection for atomic claims} 
Given an atomic claim, following \citet{gao2022attributed}, we first prompt \chatgpt to generate search queries for the claim, and then Google Search is used to get relevant web pages. 
We further split the retrieved documents into passages by sliding windows, and use a 
re-ranker combining lexical and semantic similarity to identify the most relevant passages for the given query, in which Sentence-BERT~\citep{reimers-2019-sentence-bert} serves for semantic embeddings.\footnote{\textit{cross-encoder/ms-marco-MiniLM-L-6-v2}: \url{https://huggingface.co/cross-encoder/ms-marco-MiniLM-L-6-v2}}
Finally, we aggregate evidence for all queries and select the top-5 evidences per atomic claim.

\paragraph{\factscore calculation}
\factscore~\citep{min2023factscore} is an automatic metric for fine-grained evaluation of the factuality of long-form generations. Given a generation, \factscore is calculated as the percentage of atomic claims within the generation that are supported by a knowledge source. 
For verifying the claim, we use the gathered evidences as input, along with the claim, and an instruction-tuned model as the verifier.

\paragraph{Example selection}
\figref{fig:factscoredist} shows the \factscore distribution of three component sources and the whole data set.
We keep all 45 pairs from the first source, and Dolly examples whose \factscore is less than 0.2, resulting in 33 closed question-answering pairs and 37 open questions, in total of 115 examples.
We remove a similar question (7 and 13 are similar), and four questions where the LLM did not provide helpful answers due to its inherent disability to access real-time data.
For example, the LLM cannot browse the internet and does not have access to the latest information (\textit{``which paper got the most citations in the question generation area?''} and \textit{``which large language model contains the most parameters?''}), or up-to-date data and event-specific details (``\textit{who was the general chair of COLING 2023}''), or individual information (``\textit{what are the awards that Prof. William Yang Wang have?}'').
We eventually annotated 110 examples in our first annotation stage, and more cases would be annotated in the next stage.

\subsection{Annotation} 
As many studies illustrated, annotating a LLM factuality benchmark is a highly challenging task~\citep{chen2023felm, li2023halueval}. Our preliminary trials, in which authors manually annotate labels of all subtasks, empirically confirm the pain. 

\paragraph{Preliminary Trial}
Based on the annotation guideline (see \appref{sec:guidelines}), we first conduct an in-house annotation for ten examples, each example has two annotators. We annotate the whole process for all steps and manually type results into a \textit{json} file as the pre-defined format. 
This attempt exposes three issues.

First, it is extremely time-consuming. It takes more than four hours for a fully-focused annotator to annotate a document of $\sim$400 words with about 20 sentences, in which evidence collection takes the most time and effort, particularly for topics with which the annotator is not familiar.  
Second, it is ineffective to extract relevant evidence passages by human eyes and basic string matching from retrieved Google search documents.
This not only takes time but, most importantly, takes the risk of missing the most relevant evidence due to limited traversal. It is impractical for humans to go through all relevant Web articles and select the most semantically-relevant and reliable ones in a limited time.
Humans are good at judging or making decisions, while machines are good at traversing.
Lastly. it is hard to reach a high agreement between annotators, especially for sub\-tasks of decomposition, evidence collection, and stance identification.

\subsection{Data Analysis}
\label{sec:dataanalysis}
During annotation, we remove another 16 examples because there is no standard gold answer for these questions, such as seven involving a flow of procedures, six non-factual questions, one tricky riddle-like question, one broken generated answer, and one highly-disagreed case, resulting in 94 instances.

From the perspective of LLM users, we may expect to assess any answers and identify whether they are true and reliable, including the cases deleted in our setting.
It should be highlighted that the questions involving a flow of procedures, tricky riddles, or non-factual questions need to be verified, while they are just out of the verification scope of the current fact-checking pipelines that only concern facts.
The causality and the global logic behind the whole answer are under-explored. 


\begin{figure*}[t!]
	\centering
        \includegraphics[scale=0.6]{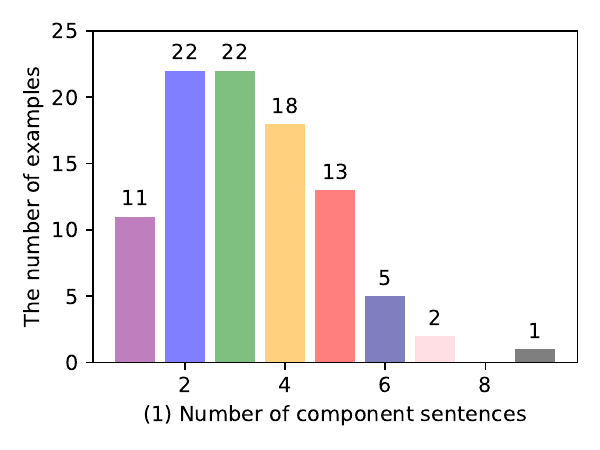} 
        \includegraphics[scale=0.6]{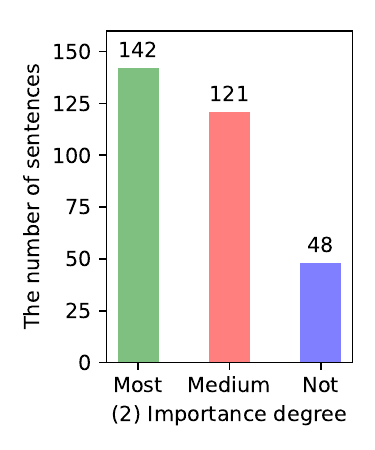}
        \includegraphics[scale=0.6]{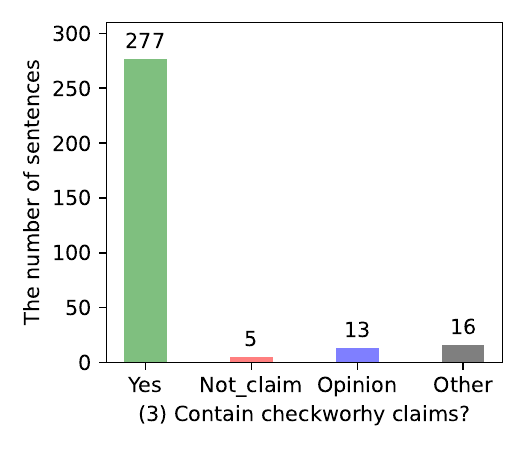}
	\caption{\textbf{Sentence analysis:} (1) Distribution of the number of sentences for each response; (2) Importance degree of sentences to answer the question (The distribution of the most important sentences to answer the question, intermediate important and not important; (3) The number of sentences across four types in terms of whether the sentence contains statements requiring fact-checking, Not\_claim refers to \textit{not a claim}, such as a question.}
	\label{fig:sent-dist}
\end{figure*}

\paragraph{Sentences:} Most responses contain 2-5 sentences, with the longest response encompassing 9 sentences as shown in \figref{fig:sent-dist}.
142 sentences are considered to be the most important sentences, 121 and 48 fall into intermediate and not important.
278 sentences contain checkworthy statements, 16, 12 and 5 are categorised into \textit{other}, \textit{opinion}, and \textit{not a claim}.

\begin{figure}[t!]
	\centering
        \includegraphics[scale=0.45]{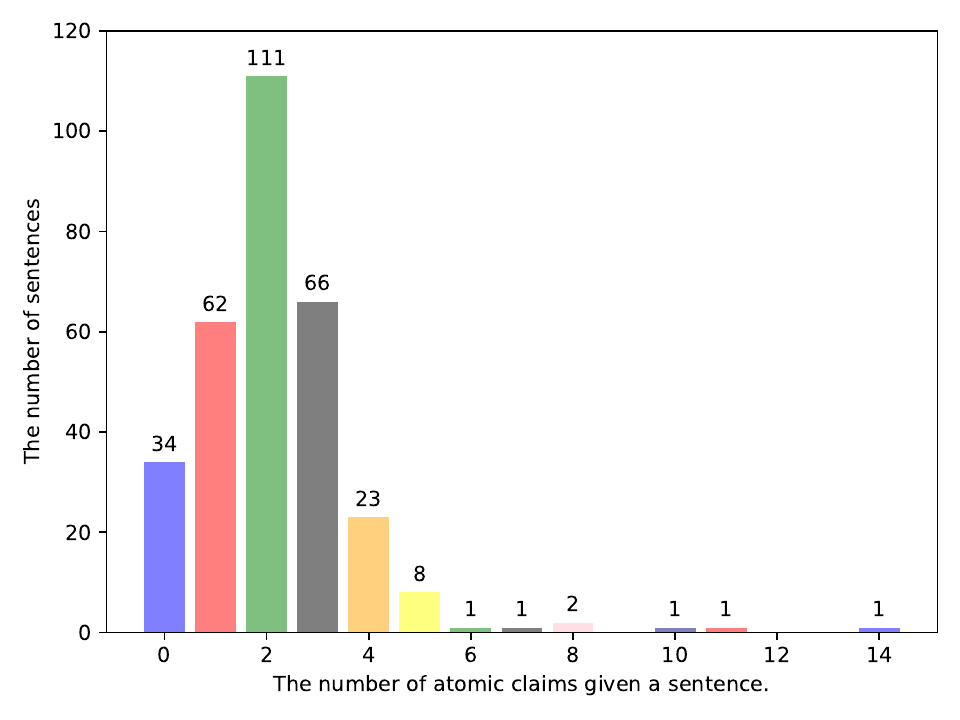}
	\caption{The distribution of component atomic claims amount given a sentence.}
	\label{fig:num-atomic-claims}
\end{figure}

\begin{figure}[t!]
	\centering
        \includegraphics[scale=0.6]{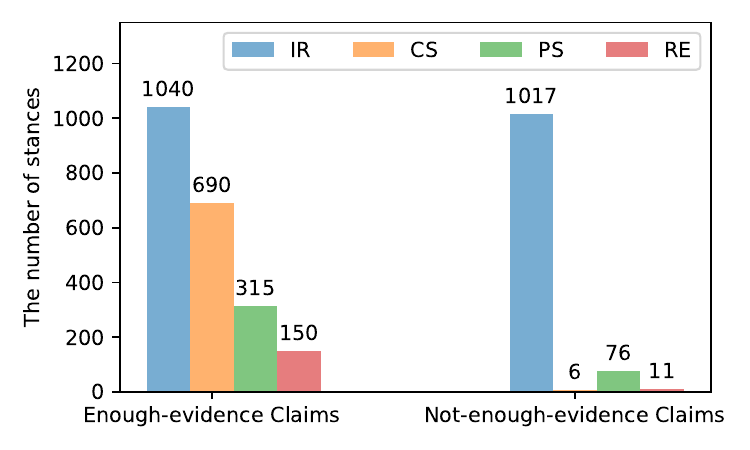}
	\caption{\textbf{Stance distribution} of claims with enough automatically-retrieved evidence to determine the factuality vs. claims without enough evidence (\textit{CP}: completely support, \textit{PS}: partially support, \textit{RE}: refute, \textit{IR}: irrelevant).}
	\label{fig:stance-dist}
\end{figure}

\paragraph{How does the evidence support the claim?}
Two-thirds pieces of irrelevant evidence (2057/3305).
We compare the stance distribution of claims in which automatically-retrieved evidence is enough to determine its factuality and the claims that cannot be determined by automatic evidence in \figref{fig:stance-dist}.
Though the majority of evidence are irrelevant for both groups, there are only 17 strong-position stances (``completely support'': \textit{CS} and ``refute'': \textit{RE}) in the latter, compared with 690 \textit{CS} and 150 \textit{RE} in the former.

\begin{figure}[t!]
	\centering
        \includegraphics[scale=0.7]{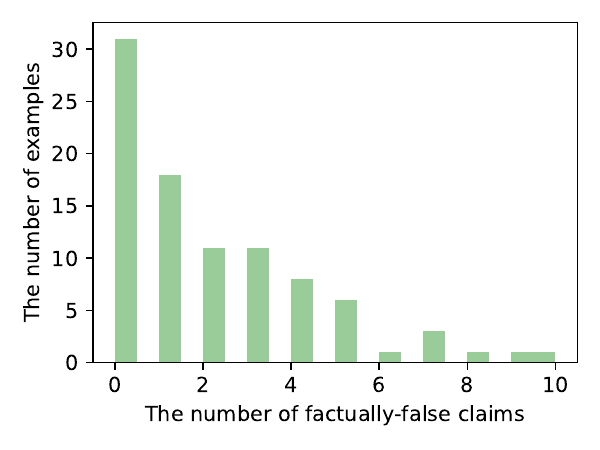}
        \includegraphics[scale=0.7]{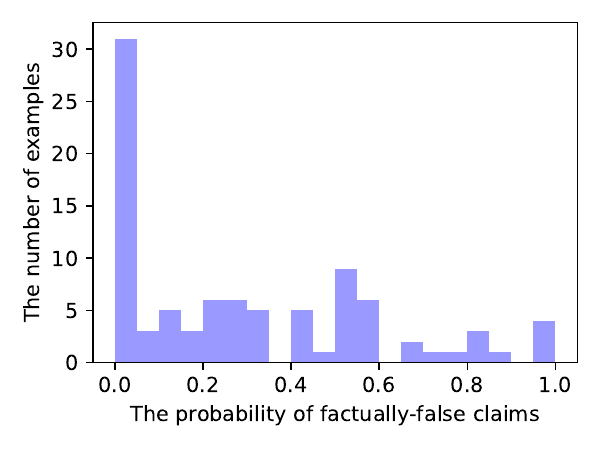}
	\caption{The number of false claims given an example.}
	\label{fig:num-false-claim-dist}
\end{figure}

\clearpage
\subsection{Prompt to Generate Atomic Claims}
\label{sec:atomicprompt}
\begin{table*}[ht!]
\centering
\caption{\textbf{Prompt} used to decompose and decontextualize a sentence into a set of independent atomic claims. We use three examples as demonstrations to elicit \chatgpt follow the instructions, break the response into sentences, as well as break a sentence into atomic claims.}
\resizebox{\textwidth}{!}{
    \begin{tabular}{p{3cm} | p{15cm}}
        \toprule
        \textbf{Field} & \textbf{Content} \\
        \midrule
        \textbf{Prompt} & 
        Depending the context, please breakdown the following sentence into independent facts. \\ &
        \\ &
        \textbf{Context:} The United States has had two black presidents: Barack Obama, who served two terms from 2009 to 2017, and Donald Trump, who served one term from 2017 to 2021. Obama was the first black president in the history of the United States. He was born in Honolulu, Hawaii, to a mother from Kansas and a father from Kenya. Trump was the second black president. He was born in New York City and previously served as a businessman and reality television personality. \\ &
        \\ &
        \textbf{The sentence is:} The United States has had two black presidents: Barack Obama, who served two terms from 2009 to 2017, and Donald Trump, who served one term from 2017 to 2021.
        \textbf{Atomic facts for this sentence are:} \\ &
        [ \\ &
            "The United States has had two black presidents: Barack Obama and Donald Trump.", \\ &
            "Black president Barack Obama served two terms from 2009 to 2017.", \\ &
            "Black president Donald Trump served one term from 2017 to 2021." \\ &
        ]\\ & 
        \\ &
        \textbf{The sentence is:} Obama was the first black president in the history of the United States.
        \textbf{Atomic facts for this sentence are:} \\ &
        [ \\ &
            "Obama was the first black president in the history of the United States." \\ &
        ]\\ &
        \\ &
        \textbf{The sentence is:} He was born in Honolulu, Hawaii, to a mother from Kansas and a father from Kenya.
        \textbf{Atomic facts for this sentence are:} \\ &
        [\\ &
            "Barack Obama was born in Honolulu, Hawaii.", \\ &
            "Barack Obama mother was from Kansas.",\\ &
            "Barack Obama father was from Kenya."\\ &
        ]\\ &
        \\ &
        \textbf{The sentence is:} Trump was the second black president.\\ &
       \textbf{ Atomic facts for this sentence are:} \\ &
        [\\ &
            "Trump was the second black president."\\ &
        ]\\ &
        \\ &
        \textbf{The sentence is:} He was born in New York City and previously served as a businessman and reality television personality. \\ &
        \textbf{Atomic facts for this sentence are:} \\ &
        [\\ &
            "Donald Trump was born in New York City.",\\ &
            "Donald Trump previously served as a businessman",\\ &
            "Donald Trump previously served as a reality television personality."\\ &
        ]\\ &
        \\
        \bottomrule
    \end{tabular}
    }
    \label{tab:atomatic-prompt-1}
\end{table*}

\begin{table*}[ht!]
\centering
\resizebox{\textwidth}{!}{
    \begin{tabular}{p{3cm} | p{15cm}}
        \toprule
        \textbf{Field} & \textbf{Content} \\
        \midrule
        &
        \textbf{Context:} In 1980, the oldest justice on the United States Supreme Court was Justice William O. Douglas. He was born on October 16, 1898, and served on the Supreme Court from 1939 until his retirement in 1975. Therefore, in 1980, Justice Douglas was still alive and would have been the oldest serving justice on the Court at that time.\\ 
        &
        \textbf{The sentence is:} In 1980, the oldest justice on the United States Supreme Court was Justice William O. Douglas.\\ &
        \textbf{Atomic facts for this sentence are:} \\ &
        [\\ &
            "In 1980, the oldest justice on the United States Supreme Court was Justice William O. Douglas."\\ &
        ] \\ &
        \\
        &
        \textbf{The sentence is:} He was born on October 16, 1898, and served on the Supreme Court from 1939 until his retirement in 1975.\\ &
        \textbf{Atomic facts for this sentence are:} \\ &
        [\\ &
            "Justice William O. Douglas was born on October 16, 1898."\\ &
            "Justice William O. Douglas served on the Supreme Court from 1939 until his retirement in 1975."\\ &
        ] \\ &
        \\ &
        \textbf{The sentence is:} Therefore, in 1980, Justice Douglas was still alive and would have been the oldest serving justice on the Court at that time.\\ &
        \textbf{Atomic facts for this sentence are:} \\ &
        [\\ &
            "Therefore, in 1980, Justice Douglas was still alive."\\ &
            "Justice William O. Douglas would have been the oldest serving justice on the Court in 1980."\\ &
        ]\\ &
        \\ &
        \textbf{Context:} There have been only four female presidents of the United States in the country's history, so it is difficult to determine an average height for this group. The four female presidents were: \\ &
        1.Abigail Adams (1797-1801) \\ &
        2.Marilyn Carlson Nelson (2009-2013) \\ &
        3.Luci Baines Johnson (1973-1977) \\ &
        4.Hillary Clinton (2017-2021) \\ &
        \textbf{The sentence is:} There have been only four female presidents of the United States in the country's history, so it is difficult to determine an average height for this group.\\ &
        \textbf{Atomic facts for this sentence are:} \\ &
        [\\ &
            "There have been only four female presidents of the United States in the country's history.",\\ &
            "It is difficult to determine an average height for four female presidents of the United States."\\ &
        ]\\ &
        \\ &
        \textbf{The sentence is:} The four female presidents were: 1.Abigail Adams (1797-1801) 2.Marilyn Carlson Nelson (2009-2013) 3.Luci Baines Johnson (1973-1977) 4.Hillary Clinton (2017-2021) \\ &
        \textbf{Atomic facts for this sentence are:} \\ &
        [\\ &
            "Abigail Adams (1797-1801) is a female president of the United States.",\\ &
            "Marilyn Carlson Nelson (2009-2013) is a female president of the United States.",\\ &
            "Luci Baines Johnson (1973-1977) is a female president of the United States.",\\ &
            "Hillary Clinton (2017-2021) is a female president of the United States."\\ &
        ]\\ &
        \\ &
        \textbf{Context:} \{\}\\ &
        \textbf{The sentence is:} \{\}\\ &
        \textbf{Atomic facts for this sentence are:}
        \\
        \bottomrule
    \end{tabular}
    }
    \label{tab:atomatic-prompt-2}
\end{table*}

\begin{table*}[t!]
\centering
\resizebox{\textwidth}{!}{
    \begin{tabular}{p{3cm} | p{15cm}}
        \toprule
        \textbf{Field} & \textbf{Content} \\
        \midrule
        Input & 
        \textbf{Context:} It is difficult to say exactly how many stars are formed in the Milky Way each year, as the rate of star formation can vary over time. However, it is generally thought that the Milky Way produces about one new star per year on average. This may seem like a small number, but keep in mind that the Milky Way is a vast galaxy with billions of stars, so even a small number of new stars added each year can add up quickly.\\ &
        \textbf{First sentence of the document:} It is difficult to say exactly how many stars are formed in the Milky Way each year, as the rate of star formation can vary over time. \\
        \midrule
        Output & 
        \textbf{Atomic facts for this sentence are:} \\ &
        [\\ &
            "It is difficult to say exactly how many stars are formed in the Milky Way each year.,",\\ &
            "The rate of star formation in the Milky Way can vary over time."\\ &
        ]\\ &
        \\ &
        \textbf{The sentence is:} However, it is generally thought that the Milky Way produces about one new star per year on average. \\ &
        \textbf{Atomic facts for this sentence are:} \\ &
        [\\ &
            "The Milky Way produces about one new star per year on average."\\ &
        ]\\ &
        \\ &
        \textbf{The sentence is:} However, This may seem like a small number, but keep in mind that the Milky Way is a vast galaxy with billions of stars, so even a small number of new stars added each year can add up quickly. \\ &
        \textbf{Atomic facts for this sentence are:} \\ &
        [\\ &
            "The Milky Way is a vast galaxy with billions of stars.",\\ &
            "Even a small number of new stars added each year can add up quickly in the Milky Way."\\ &
        ] \\
        \bottomrule
    \end{tabular}
    }
    \label{tab:atomatic-prompt-3}
\end{table*}

\clearpage
\section{Annotation Guidelines}
\label{sec:guidelines}
Based on the annotation framework, we aim to construct a benchmark for evaluation, assessing whether a fact-checking pipeline is effective in detecting and correcting factual errors in texts.
The benchmark comprises annotations for the whole fact-checking process spanning seven sub\-tasks for 110 (question, response) pairs, in which most responses are generated by \chatgpt, and some are by \gptfour.
This section introduces the annotation guidelines, and \secref{sec:dataset} and \ref{sec:annotation} provide details of data collection and annotation.

For each example, annotators are given a pair of (question, response).
A response is either an answer generated by LLMs responding to users' question, or a document returned by LLMs according to users' request.
Annotators are required to give outputs of each step shown in \figref{fig:pipeline}. 
We describe how to annotate for component sub\-tasks throughout the pipeline, particularly clarifying how to deal with possible ambiguous scenarios.

\subsection{Decompose}
It is subjective to decide the granularity of decomposition.
We may aim to break down a long document into a set of atomic claims, while the definition of an atomic claim varies.
Here, we practically apply the following strategy: 
\begin{itemize}
    \item Start by decomposing into single sentences.
    \item If the sentence contains too much information, break it into several components, but annotators do not overdo it, e.g., decomposing \textit{Capitol Hill riots happened on January 6, 2021} to one claim for a year and one for the day. 
    \item If several pieces of information are strongly dependent on each other, they are expected to co-occur in one snippet of evidence text, no more breaking-down is needed.
\end{itemize}

\subsection{De\-contextualise}
The criteria of de\-contextualisation are to ensure that all separated statements fully preserve semantics presented in the original context.
For example, a statement that \textit{it happened on Jan 6, 2021} loses information in decomposition, which makes it uncheckable. 
In such cases, annotators should replace pronouns, such as \textit{it, they, those, these, this, that}, with specific entities or events after decomposition.
De\-contextulisation is mostly needed over cases with co\-reference relation.
For complex relations, such as two sentences are strongly dependent on each other, we encourage to go back to the step of decomposition and keep the original text without breaking-down.

\subsection{Identify check\-worthy claim}
We consider two aspects in check-worthiness identification:
\begin{itemize}
    \item If a statement presents subjective opinions, then it is not checkworthy. 
    \item If the objective facts presented in a statement are commonsense, as obvious as \textit{sun rises from the east}, it is not worth checking. 
\end{itemize}
Therefore, we regard a statement as checkworthy claim when it presents objective facts, and these facts are not apparent commonsense.
There is a special case. If the objective facts presented in a statement are not publicly available information. Namely, we cannot collect any evidence over web pages related to the claim, such as personal experience.
They are regarded as uncheckable claims.

Specifically, for each statement, annotators are asked to answer two questions. 
Which category does this claim fall into? (1) factual claim; (2) subjective opinion; (3) not a claim; and (4) other.
Is this statement worth checking? (1) Yes; and (2) No.

\subsection{Retrieve and collect evidence}
Given a check\-worthy claim, annotators are asked to search and collect the five most relevant snippets of text as evidence based on general web pages (including Wikipedia pages).
Annotators are allowed to use any form of queries in retrieval, e.g.\ questions covering some aspects of the claim, or entities in the claim, and they need to record all queries and indicate those used for searching the most relevant evidence.

Note that five pieces of evidence is not a hard criterion.
If less than five (even only one) pieces of evidence are sufficient to verify the input claim, and they are from reliable sources, annotators are allowed to collect <5 results.
Meanwhile, if a claim involves a controversial topic, annotators are also encouraged to collect more than five results.

For each piece of evidence, record meta-data including (1) corresponding query, (2) citation (URL) of the web page from which this piece of evidence is extracted, (3) judgement of whether the source of evidence is reliable or not,\footnote{Source reliability can also automatically be collected from MBFC/AllSides/Politifact/, but they apply for a small number of sources.} and (4) indicator whether this individual evidence is sufficient to verify the input claim.

The aforementioned guidelines are applicable to claims for which there exists evidence over web pages.
However, there are situations where there is not any information on public web pages, e.g.\ personal experience.
They are objective facts, but are not extensively known by the public.
Put differently, they are uncheckable.
Annotators can give empty list of evidence for uncheckable claims.

\subsection{Identify evidence stance}
Given a claim and five pieces of most relevant evidence, annotators judge whether the evidence supports, partially supports, refutes or is irrelevant to the claim (see definition of stance in \secref{sec:framework}).

\subsection{Determine correction}
For a claim, there will be K snippets of text (evidence), corresponding stance vectors $[\mathbf{s}_1, \mathbf{s}_2, \dots, \mathbf{s}_K]$ and source reliability values $[r_1, r_2, \dots, r_K]$.
We skip all irrelevant evidence and follow the criteria below to determine whether edits are needed for a claim.
\begin{itemize}
    \item If the claim is completely supported by evidence, no edit.
    \item If the claim is completely refuted by evidence, check the evidence and make edits accordingly one by one.
    \item If some evidence supports the claim and some refute it, this means there are conflicts between evidence.
    In such a scenario, we consider both the source reliability and the number of evidence falling into each stance.
    If the voice of ``refute'' is stronger than ``support'', we edit, otherwise remains the original text.
    \item If some refute and some partially support, there are two possible situations depending on whether the supported partition is the same as the refuted partition: (1) if what is supported and what is refuted are the same partition, there are conflicts between evidence, follow the steps above; and (2) if they support and refute different partition of the claim, edit the refuted partition.
\end{itemize}


\subsection{Edit, Merge and Deduplicate}
In correction, we keep the principle of making minimal edits against the original text to correct factual errors. 
Annotators do not add extra information provided by evidence that is not directly targeted at factual errors.
No extra deletion, insertion or addition.
Finally, annotators merge all statements, either revised or original ones, in order, and deduplicate repeated information with the principle of minimal edits.

\clearpage
\section{Conflicting Evidence Example}
\begin{figure*}[ht!]
	\centering
        \includegraphics[scale=0.6]{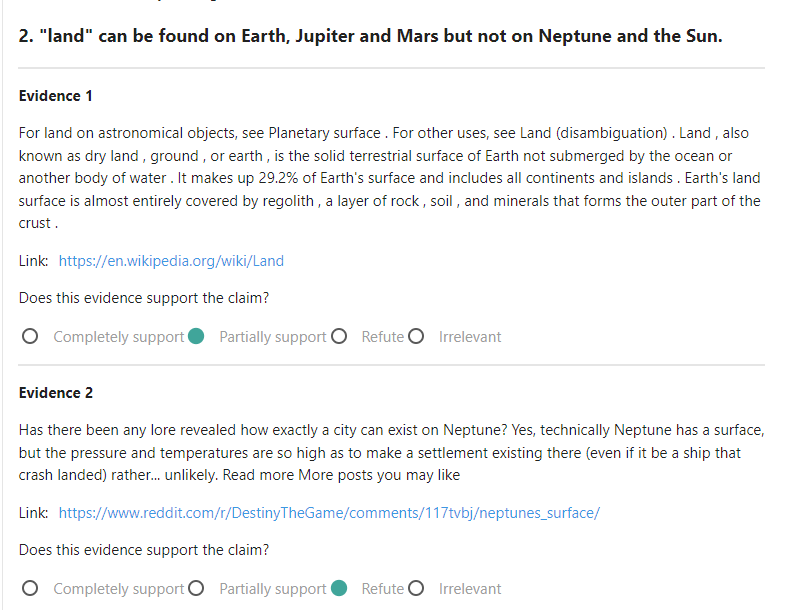}
	\caption{A claim with conflicting stance evidence: \textit{partially support} and \textit{refute}.}
	\label{fig:conflicting-evidence-example}
\end{figure*}

\clearpage
\section{Confusion Matrix of Subtasks}
\begin{figure}[ht!]
	\centering
        \includegraphics[scale=0.5]{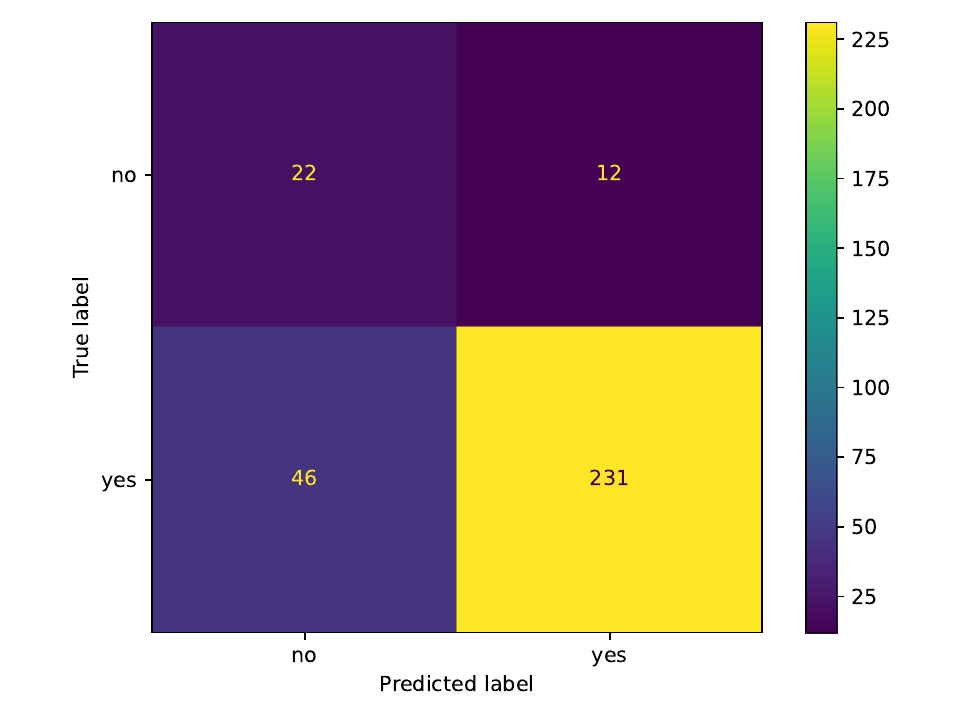} 
        \includegraphics[scale=0.5]{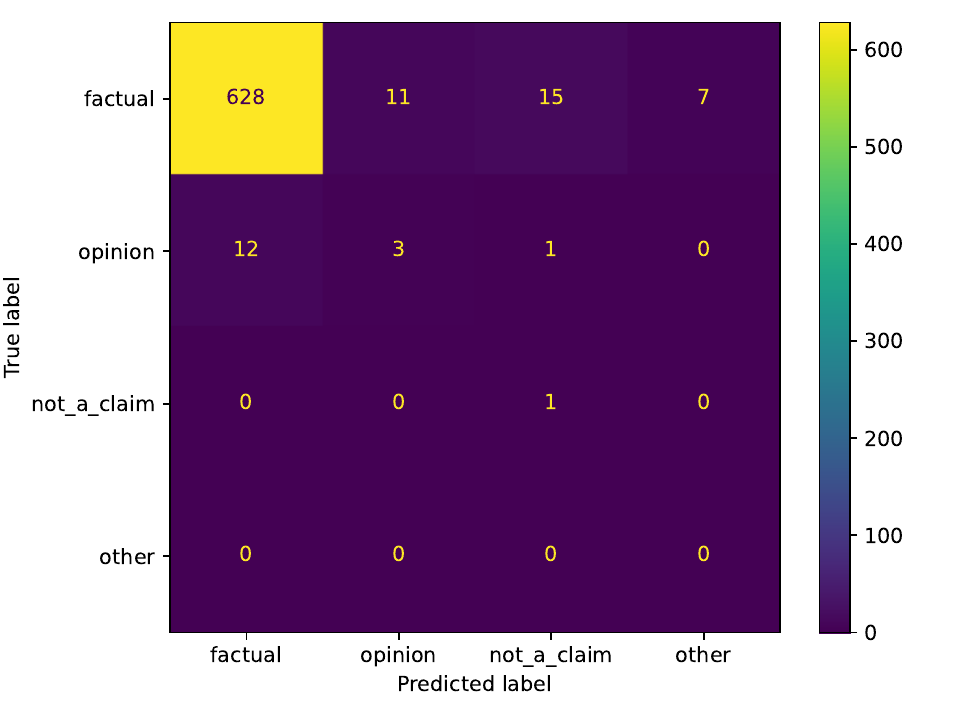} 
	\caption{\textbf{\chatgpt checkworthiness} detection confusion matrix: sentence (top) and claim (bottom)}
	\label{fig:checkworthy-cm}
\end{figure}

\begin{figure}[ht!]
	\centering
        \includegraphics[scale=0.5]{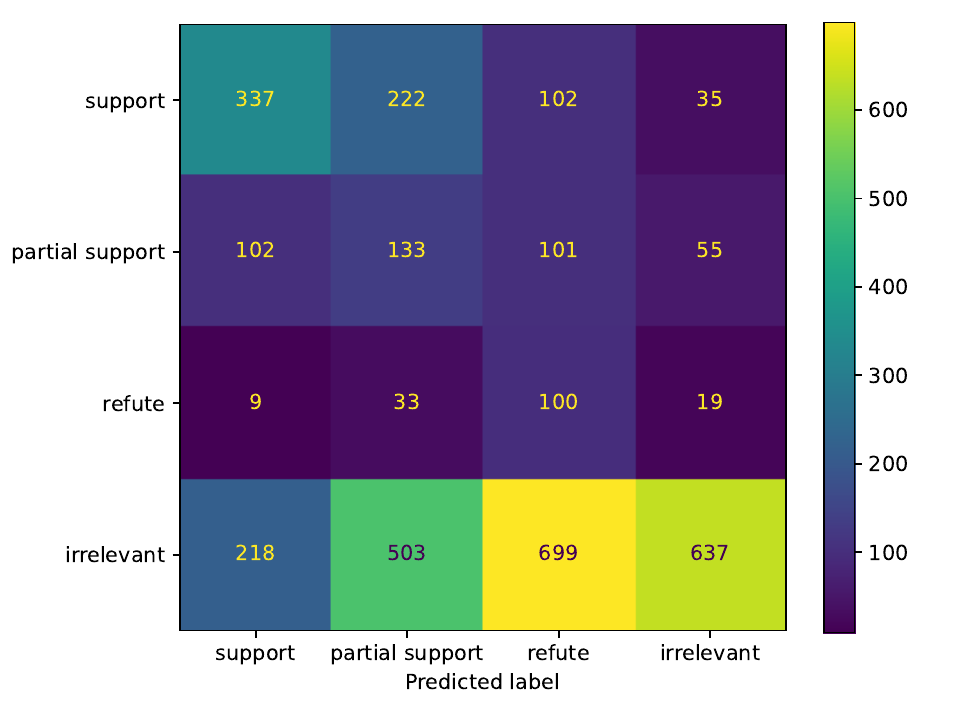} 
	\caption{\textbf{(claim, evidence) Stance} detection confusion matrix based on \chatgpt with four labels: completely support (support in short in the figure), partial support, refute and irrelevant.}
	\label{fig:stance-cm}
\end{figure}

\begin{table}[ht!]
\centering
\resizebox{0.7\columnwidth}{!}{
    \begin{tabular}{l c c c | c c c | c c c}
        \toprule
        & \multicolumn{3}{c}{\textbf{\llamatwo}} & \multicolumn{3}{c}{\textbf{\chatgpt}} & \multicolumn{3}{c}{\textbf{RoBERTa-large-mnli}}\\
        label       & Prec & Recall & F1 & Prec & Recall & F1 & Prec & Recall & F1 \\
        \midrule
        irrelevant  & 0.69 & 0.54 & 0.61 & 0.85 & 0.47 & 0.60 & 0.71 & 0.52 & 0.60\\
        refute      & 0.06 & 0.49 & 0.10 & 0.14 & 0.53 & 0.22 & 0.15 & 0.65 & 0.24\\
        support     & 0.48 & 0.12 & 0.19 & 0.53 & 0.76 & 0.63 & 0.75 & 0.65 & 0.69\\
        \bottomrule
    \end{tabular}
    }
    \caption{\textbf{Stance} detection by prompting zero-shot \chatgpt, \llamatwo and employing a fine-tuned NLI model in three label space: support, refute and irrelevant.}
    \label{tab:subtask3-results-label-specific}
\end{table}

\clearpage
\section{Example of Subtask 5}
\label{sec:subtask5-prompt}
\begin{table*}[ht!]
\centering
\resizebox{\textwidth}{!}{
    \begin{tabular}{p{3cm} | p{15cm}}
        \toprule
        \textbf{Prompt1 without question} & Given a document containing factual errors, please correct the errors in the document depending on a corresponding list of factually true claims. Note that preserve the linguistic features and style of the original document, just correct factual errors. \\
        & document: \{response\}\\
        & true claims: \{correct claims\}\\
        & revised document: \\
        & \\
        \textbf{Prompt2 with question} & Given a question, and an answer containing factual errors, please correct the errors in the document depending on a corresponding list of factually true claims. Note that preserve the linguistic features and style of the original document, just  correct factual errors. \\
        & question: \{prompt\} \\
        & document: \{response\} \\
        & true claims: \{claims\} \\
        & revised document: \\
        \midrule
        \textbf{Question} & Who was the oldest justice on the US supreme court in 1980?\\
        \textbf{Response} & In 1980, the oldest justice on the United States Supreme Court was Justice William O. Douglas. He was born on October 16, 1898, and served on the Supreme Court from 1939 until his retirement in 1975. Therefore, in 1980, Justice Douglas was still alive and would have been the oldest serving justice on the Court at that time.\\
        & \\
        \textbf{Correct claims} & In 1980, the oldest justice on the United States Supreme Court was Justice William J. Brennan Jr. \\
        & Justice William O. Douglas was born on October 16, 1898. \\
        & Justice William O. Douglas served on the United States Supreme Court from 1939 until his retirement in 1975. \\
        & In 1980, Justice William O. Douglas died. \\
        & Justice William O. Douglas was the oldest serving justice in the history United States Supreme Courtt, sat on the Court from 1939 to 1975 and died in 1980. \\
        & \\
        \textbf{Human revision} & In 1980, the oldest justice on the United States Supreme Court was Justice William J. Brennan Jr. \\
        \midrule
        \textbf{Prompt1, GPT-3.5-turbo} & In 1980, the oldest justice on the United States Supreme Court was Justice William J. Brennan Jr. He was born on October 16, 1898, and served on the Supreme Court from 1939 until his retirement in 1975. Therefore, in 1980, Justice Douglas had already passed away and would not have been the oldest serving justice on the Court at that time.\\
        & \\
        \textbf{Prompt1, GPT-4} & In 1980, the oldest justice on the United States Supreme Court was not Justice William O. Douglas, but rather Justice William J. Brennan Jr. Justice Douglas, who was born on October 16, 1898, notably served on the Supreme Court from 1939 until his retirement in 1975. However, by 1980, Justice Douglas was no longer alive, having passed away that year. Previously, he had held record as the oldest serving justice in the history of the United States Supreme Court. \\
        \midrule 
        \textbf{Prompt2, GPT-3.5-turbo} & In 1980, the oldest justice on the United States Supreme Court was Justice William J. Brennan Jr. He was born on October 16, 1898 and served on the Supreme Court from 1939 until his retirement in 1975. Therefore, in 1980, Justice Douglas had already passed away and Justice Brennan would have been the oldest serving justice on the Court at that time.\\
        & \\
        \textbf{Prompt2, GPT-4} & In 1980, the oldest justice on the United States Supreme Court was Justice William J. Brennan Jr., not Justice William O. Douglas. While Justice Douglas was indeed born on October 16, 1898, and served on the Supreme Court from 1939 until his retirement in 1975, in 1980, he was no longer alive, contrary to the original claim. It's true that Justice Douglas was considered the oldest serving justice in the history of the United States Supreme Court, his tenure spanning from 1939 to 1975, but by 1980, he had passed away. Thus, the distinction of being the oldest justice on the Supreme Court in 1980 goes to Justice Brennan Jr. \\
        \bottomrule
    \end{tabular}
    }
    \caption{Outputs (revised response) comparison using different prompts and models (GPT-3.5-turbo and GPT-4). The response by Prompt1 using GPT-4 is preferred.}
    \label{tab:subtask5-example}
\end{table*}

\clearpage
\section{Annotation Interfaces}
\label{sec:annotationinterfaces}
\subsection{Decomposition, Decontextualization and Check-worthiness detection}
\begin{figure*}[ht]
	\centering
	\includegraphics[scale=0.4]{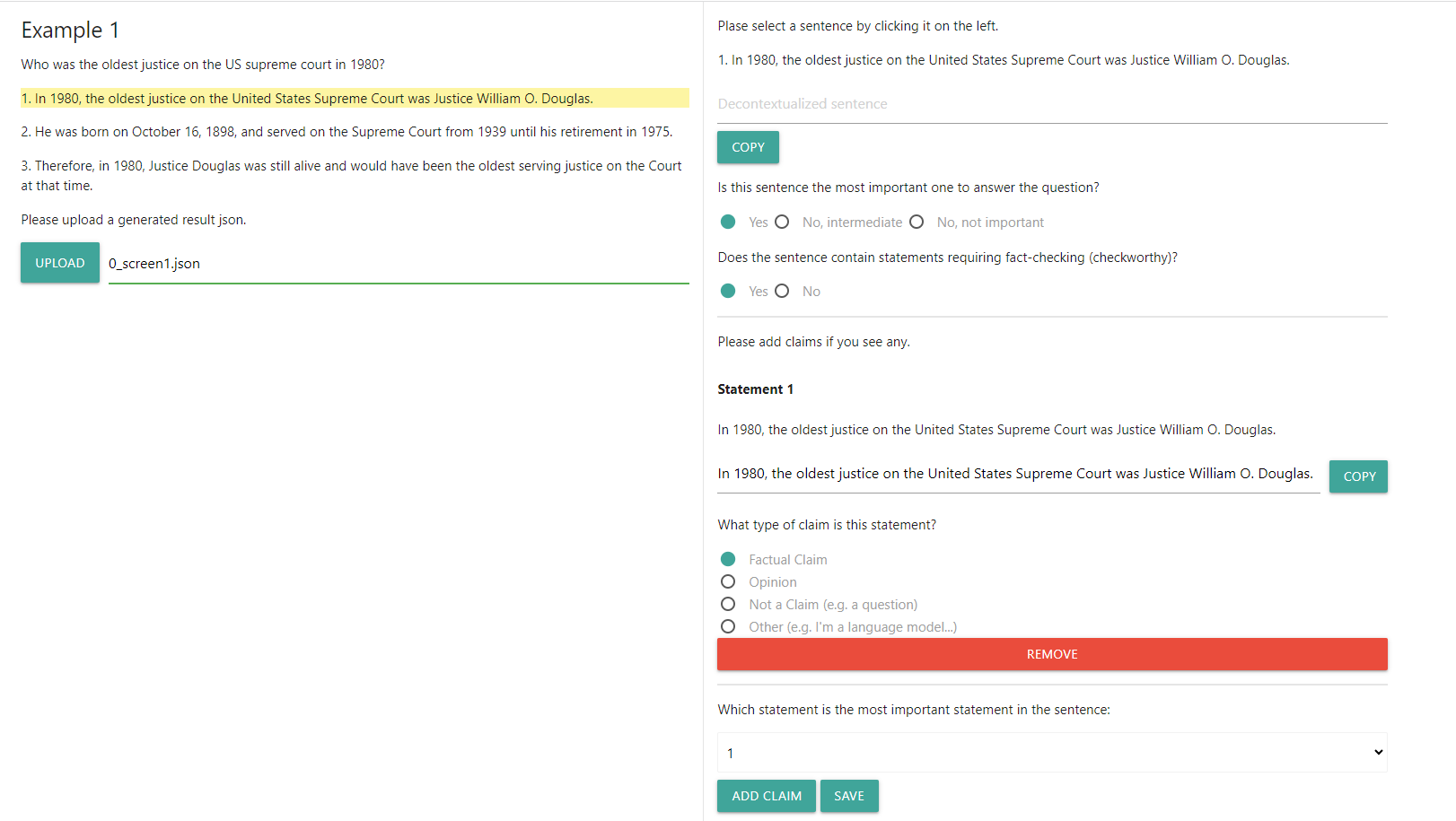} 
	\caption{Screenshot of the first annotation interface for Decomposition, Decontextualization, and Check-worthiness detection.}
	\label{fig:interface1}
\end{figure*}

\clearpage
\subsection{Evidence stance identification and Claim correction}
\begin{figure*}[ht]
	\centering
        \includegraphics[scale=0.4]{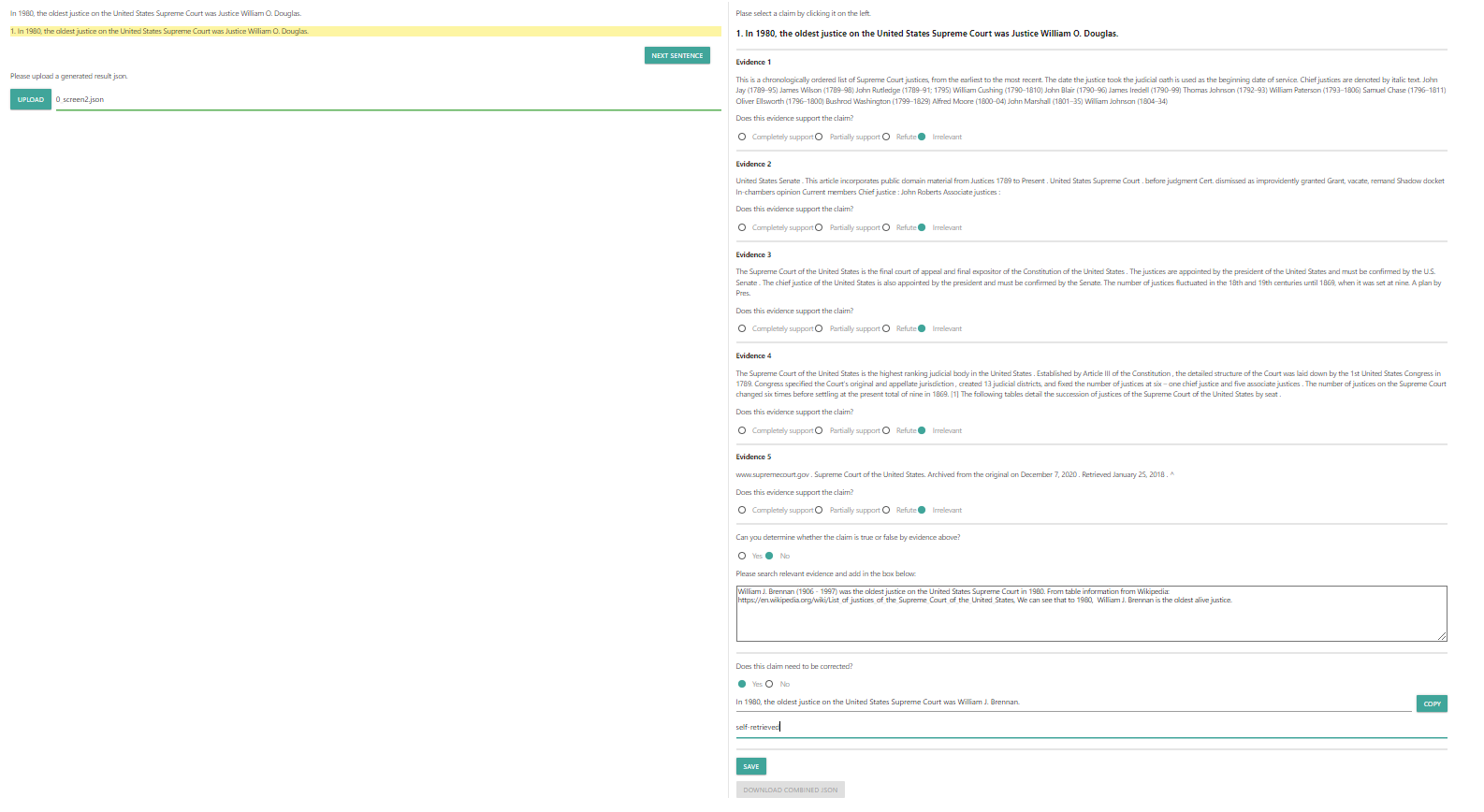}
	\caption{Screenshot of the second annotation interface: Evidence stance identification and Claim correction.}
	\label{fig:interface2}
\end{figure*}

\clearpage
\subsection{Claim Merge and De\-duplication}
\begin{figure*}[ht]
	\centering
        \includegraphics[scale=0.4]{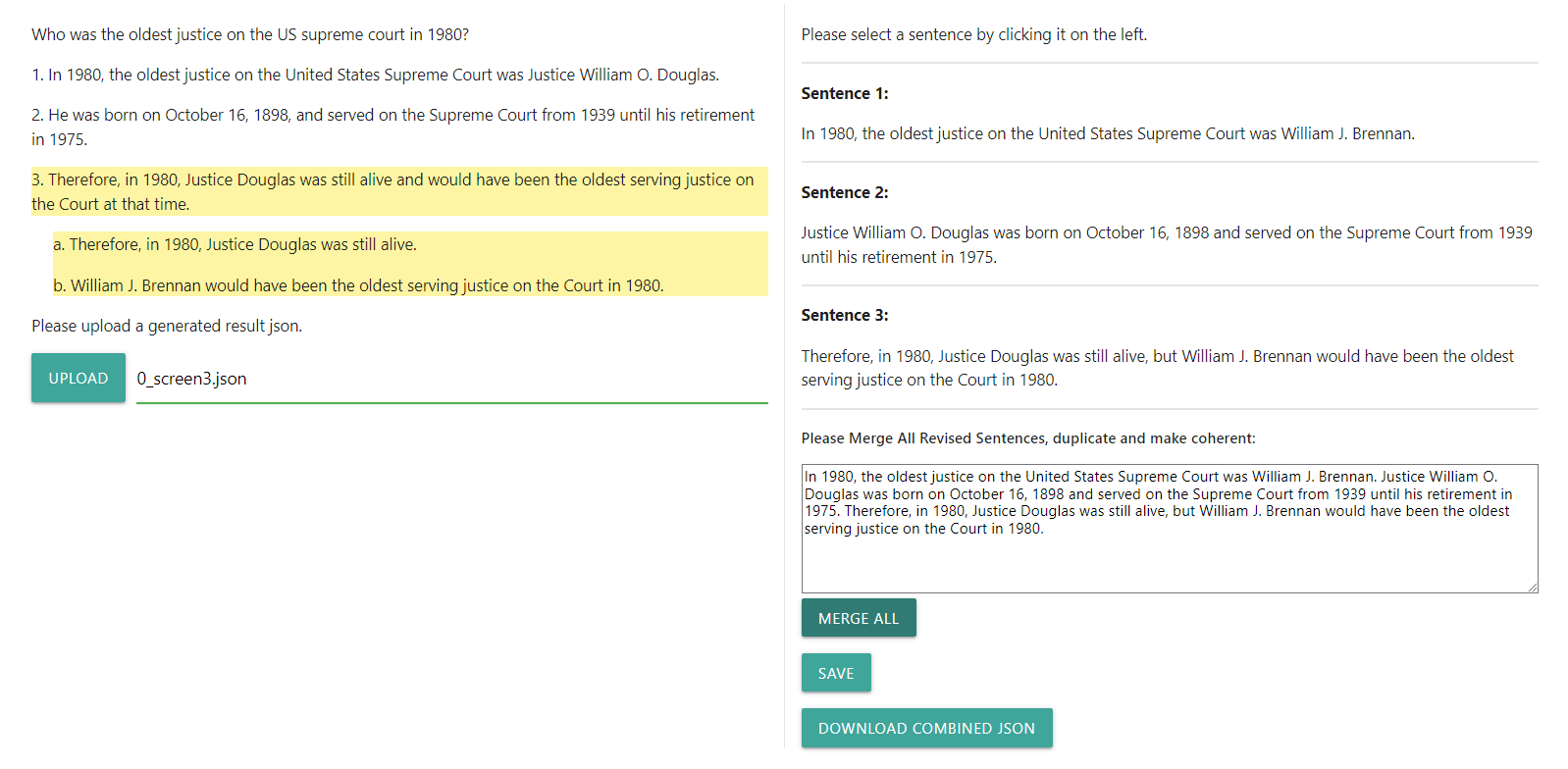}
	\caption{Screenshot of the third annotation interface: Claim Merge and De\-duplication.}
	\label{fig:interface3}
\end{figure*}



\end{document}